\documentclass{article}
\usepackage{multicol}
\usepackage{multirow}
\usepackage{booktabs}
\usepackage{graphicx}
\usepackage{algorithm}
\usepackage{array}
\usepackage{longtable}
\usepackage{algpseudocode}
\usepackage{wrapfig}
\usepackage{lipsum}
\usepackage{subcaption}
\usepackage{appendix}
\usepackage{pdflscape}
\usepackage[dvipsnames]{xcolor}

\usepackage{amsmath,amsfonts,bm}

\def\eqref#1{equation~\ref{#1}}

\def\1{\bm{1}}

\DeclareMathAlphabet{\mathsfit}{\encodingdefault}{\sfdefault}{m}{sl}
\SetMathAlphabet{\mathsfit}{bold}{\encodingdefault}{\sfdefault}{bx}{n}

\DeclareMathOperator*{\argmax}{arg\,max}

\usepackage[endLComment=,italicComments=false]{algpseudocodex}
\usepackage{tcolorbox}
\usepackage{times} %
\usepackage{utfsym}

\newcommand{\yes}{\large \color{OliveGreen}\checkmark}
\newcommand{\no}{\color{BrickRed} \scalebox{1}{\usym{2613}}}

\tcbuselibrary{breakable, skins}

\newtcolorbox{promptbox}[1]{
  title=#1,
  fonttitle=\bfseries,
  colback=black!6,
  colframe=black!30,
  colbacktitle=black!20,
  coltitle=black,
  boxrule=0.5pt,
  titlerule=0.5pt,
  arc=3pt,
  left=8pt, right=8pt, top=6pt, bottom=6pt,
  breakable
}

\definecolor{prompt1only}{RGB}{70, 130, 180}   %
\definecolor{prompt2only}{RGB}{200, 80, 80}    %

\newcommand{\pone}[1]{\textcolor{prompt1only}{#1}}
\newcommand{\ptwo}[1]{\textcolor{prompt2only}{#1}}

\usepackage[preprint]{corl_2026} %

\definecolor{softred}{HTML}{F54254}
\definecolor{softorange}{HTML}{FFB135}
\definecolor{softgreen}{HTML}{10BD35}
\definecolor{softblue}{HTML}{598BE7}
\definecolor{softpurple}{HTML}{9A1C6B}
\definecolor{plgray}{HTML}{999999}

\newcommand{\method}{GUILD}

\title{GUILD: Guided Iterative Data Collection for Imitation Learning}

\author{
  Shivin Dass\\
  University of Texas at Austin\\
  \And
  Kevin Nguyen$^\ast$\\
  University of Texas at Austin\\
  \And
  Rohan Yelandur$^\ast$\\
  University of Texas at Austin\\
  \And
  Junhong Xu\\
  University of Texas at Austin\\
  \And
  Roberto Martín-Martín\\
  University of Texas at Austin
}

\usepackage{hyperref}
\hypersetup{
  colorlinks=true,
  urlcolor=orange,
  linkcolor=orange,
  citecolor=orange,
  filecolor=orange
}

\begin{document}
\maketitle

\begin{abstract}
    Imitation learning policies in robotics are trained on datasets collected through costly real-world teleoperation. Often, the data-collection process is unguided, relying on the teleoperator's intuition, resulting in data that contributes little to overall policy performance. Recent work has shown that curating such unguided datasets can substantially improve policy performance, but these approaches operate by filtering existing datasets after collection and therefore do not prevent the acquisition of uninformative data. In this work, we propose \method{}, a novel Guided Data Collection~(GDC) method that actively reasons about which data should be collected to maximize downstream policy performance. We introduce an iterative framework that alternates between 1)~task proposal and data collection and 2)~data evaluation to close the loop. For task proposal, we leverage semantic priors from vision–language models to identify relevant environments, object configurations, and goals for data collection. However, since semantic relevance does not imply policy improvement, we perform data evaluation. By leveraging predictive data attribution we approximate how the collected trajectories influence policy performance on a target deployment setting. This signal is then used as feedback to guide subsequent data collection, closing the data collection loop. 
    We evaluate our approach in simulation and real robot settings, demonstrating that guided data collection substantially improves sample efficiency and deployment performance compared to passive data collection strategies, multiple baselines and experienced roboticists.
    More info at our \href{https://robin-lab.cs.utexas.edu/guild2026/}{website}.
\end{abstract}

\renewcommand*{\thefootnote}{\fnsymbol{footnote}}
\footnotetext{$^\ast$: denotes equal contribution. Correspondence to: \href{mailto:sdass@utexas.edu}{\texttt{sdass@utexas.edu}}\\
}


\section{Introduction}
\label{sec:intro}
Imitation learning (IL) has demonstrated a powerful paradigm for enabling robots to perform a wide range of manipulation tasks~\citep{brohan2022rt, bjorck2025gr00t, black2024pi_0, fang2026molmoact2}. 
However, successfully deploying robotic systems often requires strong performance under specific real-world conditions, where task-relevant training data may be scarce and additional data collection may be impractical or infeasible. 
Consider a robot operating in an end-user home, hospital, or warehouse: although these environments represent the ultimate deployment targets, collecting large-scale demonstrations directly within them is often constrained by privacy, safety, cost, or limited accessibility. 
In many practical settings, however, robots have access to separate training environments where interaction data is easier to acquire. 
For instance, while collecting demonstrations in the end-user's home may be difficult, humans can often gather data by teleoperating robots in other apartments, mock setups, or laboratory spaces. 
These auxiliary environments provide scalable sources of interaction, but they remain imperfectly aligned with the target domain. 
As a result, the central challenge is not simply collecting more data, but deciding what data should be collected so that it transfers most effectively under distribution shift.

A common strategy in robot learning is to collect data broadly across diverse tasks and environments~\cite{o2024open, khazatsky2024droid, walke2023bridgedata}, under the assumption that scale alone will produce robust policies. 
However, recent work in machine learning~\cite{engstrom2024dsdm, engstrom2025optimizing, li2024datacomp} and robotics~\cite{dass2025datamil, hejna2025remix, agia2025cupid} has shown that carefully curated datasets can substantially outperform much larger unfiltered ones when targeting specific downstream domains. 
These findings suggest that not all demonstrations contribute equally to policy learning. 
Existing approaches to data curation, however, primarily operate as a post-hoc filtering mechanism: they identify useful subsets only after large datasets have already been collected~\cite{dass2025datamil, agia2025cupid, memmel2025strap}.
This process is inherently inefficient, since the cost of collecting discarded data has already been incurred.
In contrast, expert roboticists often rely on intuition to decide what interactions are most valuable to collect for improving downstream performance.
Yet such intuition is difficult to formalize, expensive to acquire, and not always reliable.
Reducing the burden of robot data collection therefore requires automated methods capable of guiding humans toward collecting the most informative demonstrations for downstream policy learning.

In this work, we study the problem of Guided Data Collection (GDC)~\cite{zha2025guiding}: rather than collecting data indiscriminately and filtering it afterward, GDC methods actively reason about what data to collect, where to collect it, and which interactions to prioritize in order to maximize downstream deployment performance. 
We focus on the practical setting where direct access to the deployment environment is unavailable, but data collection remains possible in related auxiliary environments.

To address this challenge, we formulate guided data collection as an iterative optimization problem over the data-generation process itself. 
We introduce \textbf{\method{}}~(\textbf{Gu}ided \textbf{I}terative Imitation \textbf{L}earning \textbf{D}ata Collection), a framework that tightly couples task proposal, data collection, and data evaluation within a closed feedback loop. 
Given the large combinatorial space of possible environments, object configurations, and manipulation tasks, \method{} leverages semantic priors from vision--language models to define a structured proposal distribution over candidate data-collection tasks. 
However, semantic relevance alone does not imply downstream improvement. 
To address this, \method{} uses data attribution~\cite{ilyas2022datamodels, ilyas2025magic, tutorial} to estimate how individual data points influence deployment-time performance on a small set of target-domain demonstrations, and uses this signal to iteratively reallocate the collection budget toward increasingly beneficial interactions. 
By continuously refining the data-collection process based on utility, \method{} closes the loop between data collection and downstream policy evaluation. 
We evaluate \method{} in both simulation and real-world robotic settings, demonstrating substantial improvements in sample efficiency and downstream policy performance compared to several baselines, passive data collection strategies and even expert roboticists.

\section{Related Work}
\label{sec:rw}

\textbf{Data Curation and Active Data Collection in Robotics.}
A large body of work improves policy performance through post-hoc data curation, selecting useful subsets from pre-collected datasets. Generalist methods optimize training mixtures for heterogeneous robot data~\citep{hejna2025remix, zhang2025scizor,hejna2025robot}, task-specific methods retrieve relevant trajectories for a target task~\citep{memmel2025strap,du2023behavior,lin2024flowretrieval, kumarcollage}, and attribution-based approaches estimate the influence of demonstrations on downstream performance~\citep{dass2025datamil,agia2025cupid}. Unlike these methods, which assume data have already been collected, we study how to decide what data to collect in the first place. 
 Another line of work uses Bayesian optimization to actively acquire informative data~\citep{frazier2018tutorial}, but these approaches typically scale only to low-dimensional parameter spaces and are primarily used for sim-to-real parameter tuning~\citep{rai2018bayesian,rai2019using,ramos2019bayessim,mehta2020active,muratore2021bayrn}. Our work is most closely related to Factored Scaling Curves (FSC)~\citep{zha2025guiding}, which also selects environmental factors for target deployment. However, FSC requires training multiple policies per factor making it expensive to compute, whereas \method{} estimates utility using predictive data attribution from a single policy training. Furthermore, unlike FSC, we do not assume factors are predefined, instead combining predictive attribution with VLM-based task generation to efficiently explore open-ended task spaces.

\textbf{Automatic Environment and Data Generation.}
To scale robot data collection, prior work has explored automatic generation of tasks, scenes, and demonstrations~\citep{mandlekar2023mimicgen, li2025momagen,dass2022pato}. Earlier approaches rely on procedural heuristics to compose environments and task specifications~\citep{dai2025manitaskgen,deitke2022,shukla2025maniskill}, while more recent methods use foundation models to generate manipulation tasks and synthesize demonstrations via motion planning~\citep{gong2025anytask,wang2023gensim,hua2025gensim2}. In reinforcement learning, language models have also been used to generate reward functions, domain randomization parameters, and training environments~\citep{ma2024eureka,ma2024dreureka,wang2023robogen,liang2024eurekaverse}. \method{} shares its VLM-based task proposal stage with methods such as GenSim~\citep{wang2023gensim,hua2025gensim2} and Eurekaverse~\citep{liang2024eurekaverse}, but differs fundamentally in objective. Existing methods primarily target zero-shot generalization and therefore generate broad, undirected task distributions. In contrast, \method{} guides generation toward tasks most relevant for a target deployment setting.

\textbf{Data Attribution.} 
We build on \emph{data attribution}~\citep{ilyas2022datamodels, ilyas2025magic,tutorial,koh2017understanding,hammoudeh2024training,park2023trak,chang2024scalable}, a body of work that predicts model behavior as a function of training data. 
These methods have been used to improve language-model pretraining~\citep{engstrom2024dsdm} and instruction tuning~\citep{engstrom2025optimizing,xia2024less,liu2024tsds}, to increase worst-group robustness~\citep{jain2024improving}, and to remove outliers in supervised learning~\citep{lin2022measuring}; within robotics, they have been used to curate already-collected demonstrations~\citep{dass2025datamil,agia2025cupid}.
However, these methods operate strictly as post-hoc curation mechanisms on static, already-collected offline datasets. 
In contrast, \method{} embeds data attribution into an open-ended data collection loop. 
Rather than merely filtering existing datasets, we use attribution as a signal to guide future data acquisition.

\begin{figure*}[t]
    \centering
    \includegraphics[width=1.0\linewidth]{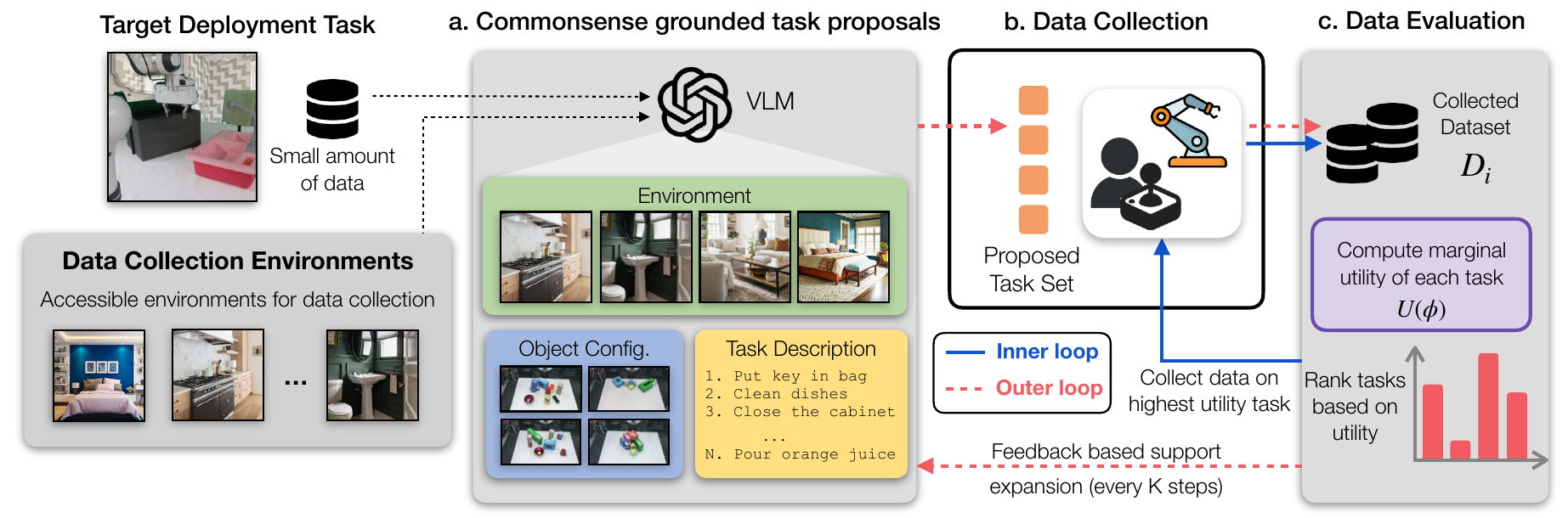}
    \caption{\textbf{Overview of \method{} Pipeline.} Given a limited-access deployment setting and a set of accessible data collection environments, \method{} uses VLM priors to propose candidate scenes, object placements and tasks for data collection. Collected trajectories are then evaluated using predictive data attribution to estimate their influence on deployment-task performance. These attribution scores are used as feedback to guide subsequent data collection and task proposals---closing the loop.}
    \label{fig:overall_pipeline}
    \vspace{-1.5em}
\end{figure*}

\section{Preliminaries}
\label{sec:preliminaries}

\textbf{Policy Learning.}
Our work focuses on {\em imitation learning}, where the goal is to learn a policy $\pi$ that 
maps states $s$ to distributions over actions $a$ using a collection of trajectories $\mathcal{D}=(\tau_1, \tau_2\cdots\tau_n)$ of state-action pairs. 
We will define a policy learning {\em algorithm} $\mathcal{A}$~(eg. stochastic gradient descent on an imitation learning objective), as
a function that takes as input a dataset $\mathcal{D}$ and outputs an optimized policy $\pi$. We measure the performance of the policy $\pi$ using a {\em metric} $\mathcal{M}: \pi \rightarrow \mathbb{R}$, which is a function mapping policies to a scalar value. The most common choice of metric is {\em success rate}---the fraction of times rolling out the policy results in the policy completing a given task---but our notation is general and can also capture other choices of $\mathcal{M}$.

\textbf{Guided Data Collection.}
In Guided Data Collection~(GDC)~\citep{zha2025guiding}, we are given (i)~an initial dataset \(\mathcal{D}\), (ii)~a learning algorithm \(\mathcal{A}\), and (iii)~a target metric \(\mathcal{M}\). The goal is to collect an additional dataset \(\mathcal{D}'\) under a fixed budget \(k=|\mathcal{D}'|\) such that training on \(\mathcal{D}\cup\mathcal{D}'\) maximizes downstream policy performance under \(\mathcal{M}\). We model data collection through a parameterized generator
\[
\tau\sim\mathcal{G}(\phi),
\]
where \(\phi\) specifies tasks (or interchangeably task parameters) such as scene configuration, object placements, and task goals.
While our setting uses human teleoperators as data collectors, the formulation also captures scripted controllers, motion planners, or autonomous policies.

Importantly, effective transfer across domains typically requires collecting data from a \textit{diverse set} of tasks and environments~\citep{o2024open, dass2025datamil, maddukuri2025sim}. Consequently, instead of optimizing a single task parameter $\phi$, we optimize a distribution over task parameters \(p_\theta(\phi)\) for data collection. Given a budget \(k\), we sample
\(
\phi\sim p_\theta(\phi)
\)
and collect a dataset \(\mathcal{D}'_\theta\). The Guided Data Collection objective is therefore:
\begin{equation}
\label{eq:optim}
\argmax_\theta\;
\mathcal{M}
\left(
\mathcal{A}
\left(
\mathcal{D}\cup\mathcal{D}'_\theta
\right)
\right).
\end{equation}
This formulation treats data collection itself as an optimization problem, where the learner actively allocates a limited collection budget toward the tasks and environments that most improve downstream deployment performance. We focus on a practical setting where the target environment is only sparsely accessible for data collection, e.g., robots deployed in end-user homes but trained on data collected in auxiliary environments such as laboratory apartments or mock kitchens. As a result, the training and deployment distributions may differ substantially. \method{} seeks to allocate the collection budget toward tasks and environments that best improve deployment-time performance under this distribution shift, assuming access to a small target dataset \(\mathcal{D}_{\mathrm{targ}}\) for evaluation.

\textbf{Data Attribution.}
Data attribution studies the problem of estimating how individual training examples influence downstream model behavior. Let
$f(\mathcal{D}) = \mathcal{M}(\mathcal{A}(\mathcal{D}))$
denote the performance of a policy trained on a dataset $\mathcal{D}$. Since directly evaluating $f(\mathcal{D})$ requires training and evaluating a policy, which is computationally expensive, data attribution methods instead seek to learn a cheaper surrogate $\widehat{f}(\mathcal{D}) \approx f(\mathcal{D})$ that predicts downstream performance as a function of the training data. 
There are several ways of computing $\widehat{f}$~\cite{ilyas2022datamodels, ilyas2025magic, park2023trak} and in Section~\ref{subsec:attribution}, we explore one such option and discuss how we use it to define the marginal utility of a task, i.e. the effect of adding an additional trajectory from a task to the downstream policy performance on the target task.

\section{\method{}: Guided Iterative Imitation Learning Data Collection}
\label{sec:method}
\begin{wrapfigure}{r}{0.48\textwidth}
\vspace{-3em}

\begin{minipage}{0.48\textwidth}

\begin{algorithm}[H]
\caption{\method{} Pseudocode}
\label{alg:method}
\begin{algorithmic}
\footnotesize

\BeginBox[fill=white]
\State \textbf{Input:} target context $c_{\mathrm{targ}}$, target dataset $\mathcal{D}_{\mathrm{targ}}$, data collector/generator $\mathcal{G}$, learner $\mathcal{A}$, metric $\widehat{\mathcal{M}}$
\State \textbf{Hyperparameters:} outer iterations $N$, inner iterations $K$, batch size $b$, number of new tasks $m$
\EndBox

\LComment{\color{softblue} Initial task proposal}
\State $\mathcal{T}_0 \gets \textsc{VLM-Propose}(c_{\mathrm{targ}}, m)$
\State $\mathcal{D} \gets \textsc{CollectUniform}(\mathcal{G}, \mathcal{T}_0)$

\LComment{\color{softblue} Outer loop}
\For{$i \gets 0$ \textbf{to} $N-1$}

    \LComment{\color{softblue} Inner loop}
    \For{$j \gets 0$ \textbf{to} $K-1$}

        \State $\mathcal{I} \gets \textsc{Attribution}(\mathcal{A}, \widehat{\mathcal{M}}, \mathcal{D}, \mathcal{D}_{\mathrm{targ}})$

        \LComment{\color{softblue} Compute marginal task utilities}
        \ForAll{$\phi \in \mathcal{T}_i$}
            \State $U(\phi) \gets
            \mathbb{E}_{\tau_k \sim \mathcal{D}_\phi}
            \left[
            \mathcal{I}_k
            \right]$
        \EndFor

        \State $\phi^* \gets \arg\max_{\phi \in \mathcal{T}_i} U(\phi)$

        \State $\mathcal{B} \gets \textsc{Collect}(\mathcal{G}, \phi^*, b)$

        \State $\mathcal{D} \gets \mathcal{D} \cup \mathcal{B}$

    \EndFor

    \LComment{\color{softblue} Feedback-driven support expansion}
    \State $\mathcal{T}_{\mathrm{new}} \gets
    \textsc{VLM-Propose}
    \big(c_{\mathrm{targ}}, \textsc{Rank}(\mathcal{T}_i,U), m\big)$

    \State $\mathcal{D} \gets \mathcal{D} \cup \textsc{CollectUniform}(\mathcal{G}, \mathcal{T}_{new})$ 
    
    \State $\mathcal{T}_{i+1} \gets \mathcal{T}_i \cup \mathcal{T}_{\mathrm{new}}$

\EndFor

\LComment{\color{softblue} Final policy training}
\State \Return final dataset $\mathcal{D}_N$ and policy $\mathcal{A}(\mathcal{D}_N)$

\end{algorithmic}
\end{algorithm}
\end{minipage}

\vspace{-2em}
\end{wrapfigure}

Directly optimizing the GDC objective in Eq.~\ref{eq:optim} is intractable due to the large, open-ended space of collection parameters \(\phi\). \method{} instead approximates the optimal collection distribution \(p_{\theta^*}(\phi)\) through an iterative closed-loop procedure that alternates between: (1)~expanding the support of the collection distribution with semantically plausible task parameters, and (2)~updating the sampling mass over this support using attribution-based utility estimates.

The first component (Section~\ref{subsec:vlm}) uses vision--language models (VLMs) to propose a candidate task set
\(
\mathcal{T}=\{\phi_1,\phi_2,\dots,\phi_n\},
\)
defining the current support of \(p_\theta(\phi)\). Rather than directly optimizing policy performance, the VLM proposes task parameters that are semantically grounded in the target domain and feasible within the available data collection environments. The second component (Section~\ref{subsec:attribution}) uses predictive data attribution to estimate the marginal utility of demonstrations collected from each task, providing a local signal for shifting the empirical collection distribution toward \(p_{\theta^*}(\phi)\).

\method{} operates at two timescales. An outer loop (indexed by $i$) periodically expands the support of the collection distribution by generating new task candidates \(\mathcal{T}_i\), while an inner loop~(indexed by $j$) performs multiple rounds of attribution-guided data collection with fixed support. During these inner-loop updates, \method{} allocates collection effort toward higher-utility tasks before expanding the support again. By alternating between support expansion, data collection, and evaluation, \method{} refines the collected data distribution toward tasks predicted to maximally improve deployment performance. An overview is shown in Figure~\ref{fig:overall_pipeline}, with pseudocode in Algorithm~\ref{alg:method}.

\subsection{Task-Parameter-Space Exploration via Commonsense-Grounded Proposals}
\label{subsec:vlm}

The space of possible task parameters \(\phi\) is too large to enumerate exhaustively. \method{} therefore uses a VLM to generate a semantically grounded candidate task set \(\mathcal{T}_i=\{\phi_1,\phi_2,\dots,\phi_n\}\), defining the current support of the collection distribution \(p_\theta(\phi)\), i.e. the tasks considered for subsequent data collection. This support-expansion stage forms the outer loop of \method{}: \(\mathcal{T}_i\) is generated once per outer-loop iteration and is fixed during several subsequent rounds of attribution-guided collection in the inner-loop.

Given target-domain context---video demonstrations, language descriptions, and available training assets (e.g., environments and objects)---the VLM proposes feasible task parameters specifying the collection environment, scene configuration, and manipulation task. The VLM does not determine how much data should be collected from each task; instead, it identifies promising regions of the open-ended task space, while attribution-based evaluation later determines their utility.

\paragraph{Initialization.}
At the first outer-loop iteration (\(i=0\)), \method{} generates an initial task set \(\mathcal{T}_0\) and uniformly samples from it to collect a seed dataset
\(
\mathcal{D}_0\sim\mathcal{G}(\mathcal{T}_0).
\)
This initialization provides broad coverage over semantically plausible tasks and establishes the initial data distribution. The resulting dataset \(\mathcal{D}_0\) serves as the starting point for attribution-based utility estimation.

\textbf{Feedback-driven support expansion.}
After every $K$ inner-loop collection iterations, \method{} expands the support of the collection distribution by querying the VLM for additional task parameters \(\mathcal{T}_{\mathrm{new}}\), updating the support as
\(
\mathcal{T}_i
=
\mathcal{T}_{i-1}
\cup
\mathcal{T}_{\mathrm{new}}.
\)
To guide this expansion, the VLM receives both the original target-domain context and a ranked list of previously evaluated task parameters derived from attribution-based utility estimates. This provides grounded feedback and biases exploration toward regions semantically related to high-utility tasks, adding support for promising new tasks outside the ones previously explored. In this way, the VLM expands the support of \(p_\theta(\phi)\), while attribution determines how collection effort is allocated within it. Similar to \emph{initialization}, we add a uniform seed dataset for $T_\text{new}$ to our dataset. We ablate the feedback-driven support expansion mechanism in Appendix~\ref{app:no_feedback} and show that including the feedback mechanism matches or improves the performance over a no-feedback control across all tasks.

\subsection{Data Evaluation via Predictive Attribution}
\label{subsec:attribution}

After each collection round, \method{} updates the collection distribution over the current task set by estimating which tasks are most likely to improve deployment-time performance if additional demonstrations are collected from them. Rather than using attribution purely for post-hoc dataset curation, \method{} interprets attribution scores as local estimates of how the empirical data distribution should shift towards the optimal collection distribution \(p_{\theta^*}(\phi)\).

\textbf{Learning attribution model.}
Recall from Section~\ref{sec:preliminaries} that data attribution methods learn a surrogate
\(
\widehat{f}(\mathcal{D}) \approx f(\mathcal{D})
\)
that predicts downstream policy performance as a function of the training dataset. \method{} uses this surrogate to estimate how small changes in the composition of the collected data affect target-domain performance. We instantiate this using MAGIC~\citep{ilyas2025magic}, which leverages metagradients to efficiently approximate the surrogate. Following~\citep{engstrom2025optimizing,ilyas2025magic}, we parameterize the dataset using weights
\(
\mathbf{w}\in[0,1]^n
\),
where \(w_k\) specifies the contribution of trajectory $\tau_k$ during training. A value of $w_k=1$ corresponds to fully including $\tau_k$ in training, while $w_k=0$ excludes it entirely. Treating \(\mathbf{w}\) as meta-parameters, a first-order Taylor expansion gives:
\begin{equation}
\label{eq:metagradient_approx}
\mathcal{M}(\mathcal{A}(\mathbf{w}))
\approx
\mathcal{M}(\mathcal{A}(\mathbf{w}_0))
+
\nabla_\mathbf{w}
\mathcal{M}(\mathcal{A}(\mathbf{w}_0))^\top
(\mathbf{w}-\mathbf{w}_0),
\end{equation}
where
\(
\mathbf{w}_0=\mathbf{1}
\)
corresponds to training on the full dataset. The gradient
$\mathcal{I}
~=~
\nabla_\mathbf{w}
\mathcal{M}(\mathcal{A}(\mathbf{w}_0))
$
measures how infinitesimal changes in trajectory weights affect downstream policy performance, with each entry \(\mathcal{I}_k\) acting as a trajectory-level attribution score for trajectory $\tau_k$. Recent work~\citep{engstrom2025optimizing} shows how $\mathcal{I}$ can be efficiently computed.
Directly evaluating \(\mathcal{M}\) requires expensive policy rollouts in the target environment. Following~\citet{dass2025datamil}, we substitute $\mathcal{M}$ in Eq.~\ref{eq:metagradient_approx} with an offline proxy metric based on the validation loss over a small target dataset:
$
\widehat{\mathcal{M}}(\pi,\mathcal{D}_{\mathrm{targ}})
=
\sum_{(s,a)\in\mathcal{D}_{\mathrm{targ}}}
\mathcal{L}(\pi(s),a),
$
which provides an efficient surrogate for learning the attribution model. 

\textbf{Marginal Task Utility.}
Since demonstrations are collected by selecting task parameters $\phi$, \method{} aggregates trajectory-level attribution scores into a task parameter-level utility.
Let $\mathcal{D}_\phi \subset \mathcal{D}$ denote the set of trajectories collected under task parameter $\phi$.
\method{} defines the marginal utility of $\phi$ as \begin{equation} 
U(\phi)
=
\frac{1}{|\mathcal{D}_\phi|}
\sum_{\tau_k \in \mathcal{D}_\phi}
\mathcal{I}_k.
\end{equation} 
Intuitively, $U(\phi)$ approximates the expected marginal benefit of collecting an additional trajectory from task parameter $\phi$, under the current dataset composition.
This utility provides a one-step approximation to the update direction for the data collection distribution towards the optimal distribution of task parameters $p_{\theta^*}(\phi)$: task parameters with higher utility should receive more sampling mass, while task parameters with low or negative utility should receive less.

\textbf{Batched Iterative Collection.}
The attribution model is local to the current dataset composition; repeatedly collecting from a high-utility task may reduce its marginal value as that region becomes sufficiently covered. \method{} therefore performs \emph{batched iterative collection}, recomputing utilities after each update. At inner-loop iteration \(j\), \method{} estimates attribution scores using the current dataset \(\mathcal{D}_j\), computes task utilities \(U(\phi)\) for each supported task parameters \(\phi\in\mathcal{T}_i\), and selects task parameters with the highest marginal utility:
\begin{equation}
\phi_j^*
\in
\arg\max_{\phi\in\mathcal{T}_i}
U(\phi).
\end{equation}
A small batch of demos \(B_j\) is then collected with $\phi_j^*$ and added to the dataset:
\(
\mathcal{D}_{j+1}
=
\mathcal{D}_j \cup B_j.
\)
The support \(\mathcal{T}_i\) remains fixed during these inner-loop updates, allowing \method{} to locally optimize allocation of collection effort over the current task set. This update shifts the empirical collection distribution toward task parameters that locally approximate $p_{\theta^*}(\phi)$ under the current attribution estimate. 
Utilities are then recomputed on the updated dataset before the next collection round. We also discuss a soft-selection strategy in Appendix~\ref{app:soft_sampling}.

The batch size controls the tradeoff between compute cost and adaptivity: 
smaller batch allow \method{} to more frequently recompute attribution scores and respond to changes in marginal utility, while larger batches reduce computation at the cost of less precise distribution updates. After $K$ inner-loop updates, \method{} returns to the support-expansion stage (Section~\ref{subsec:vlm}) to propose new task parameters beyond the current support.

\begin{figure*}[t]
    \centering
    \includegraphics[width=1.0\linewidth]{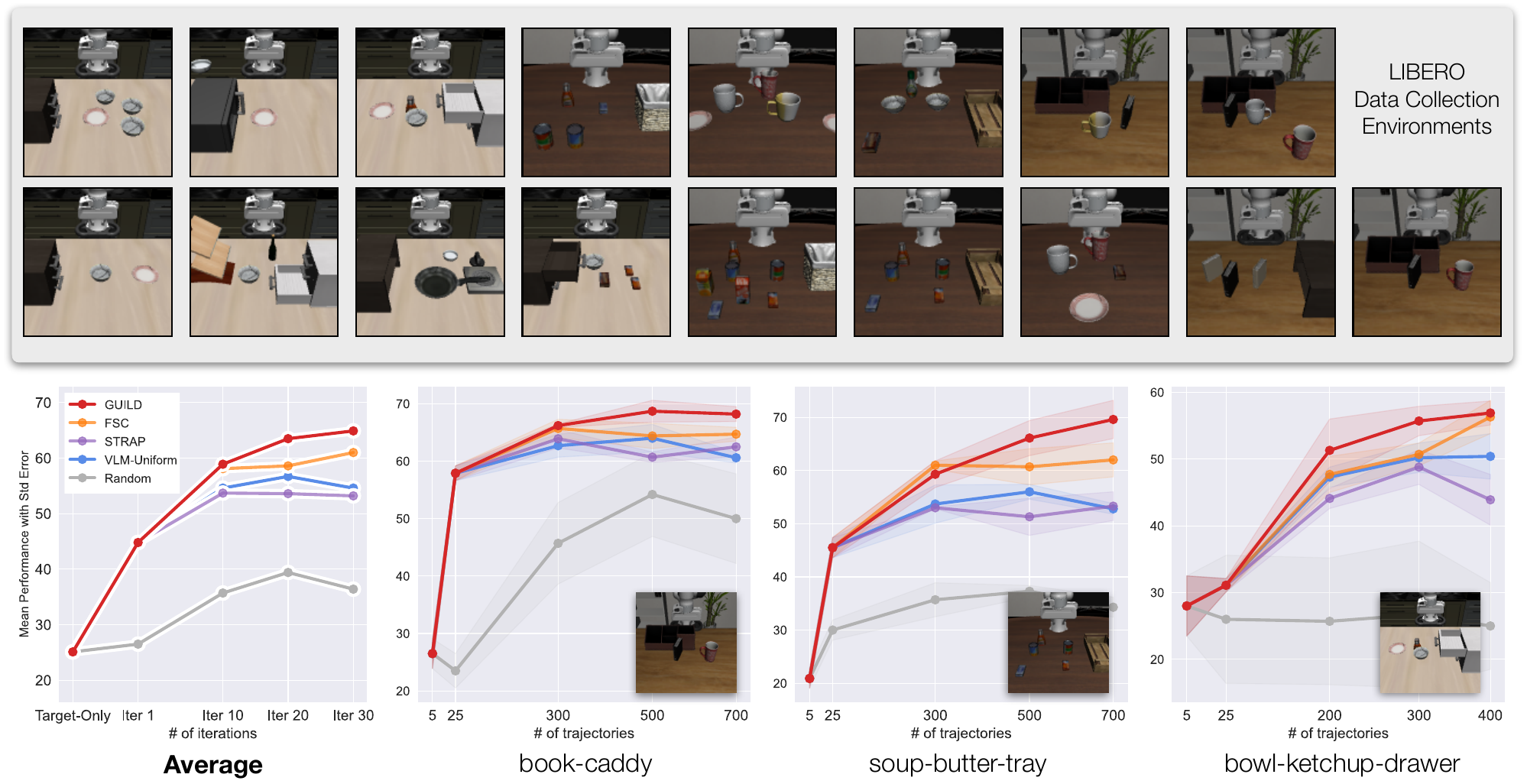}
    \caption{\textbf{LIBERO guided data collection.} We generate data using various strategies and train policies across different points of the process. \method{} either matches or surpasses the next best baseline. For each method on each task, we train 3 different data collection seeds. For every evaluation, we train 5 policies, measure their performance over 100 rollouts and average their performances. The means and standard error are computed across the 3 seeds of data collection process.}
    \label{fig:libero_res}
    \vspace{-2em}
\end{figure*}

\section{Experiments}
\label{sec:experiments}

\textbf{Training Details.} 
We train a language-conditioned Octo~\citep{octo_2023} policy on the collected datasets, initializing from the pretrained checkpoint released by the authors. Following prior work~\citep{khazatsky2024droid,dass2025datamil,lin2024flowretrieval,maddukuri2025sim}, we co-train the policy with the target-task demonstrations \(\mathcal{D}_{\mathrm{targ}}\) using a 50/50 sampling mixture.

\textbf{Baselines.}
We compare \method{} with the following baselines:
\textbf{Random}, which samples tasks by randomly selecting objects and semantically valid placements and collects data uniformly across them;
\textbf{VLM-Uniform}, \textbf{STRAP}~\citep{memmel2025strap} and Factored Scaling Curves~(\textbf{FSC})~\citep{zha2025guiding}, all use the same VLM~(GPT 5.2~\citep{singh2025openai}) \emph{task proposer} as Section~\ref{subsec:vlm}; \textbf{VLM-Uniform} samples uniformly from the proposed tasks without evaluation;
\textbf{STRAP} evaluates task relevance using DINOv2~\citep{oquab2023dinov2} embedding similarity; and
\textbf{FSC} estimates task utility by training policies on varying amounts of task-specific data and measuring the resulting scaling behavior. See Appendix~\ref{app:baselines} for implementation details.

\subsection{Results}
\emph{Can \textbf{\method{}} efficiently guide data collection?}

We first evaluate \method{} on LIBERO, using 17 environments from LIBERO-90 (Figure~\ref{fig:libero_res}, top). We select three target tasks from distinct scenes and, for each target, collect training data exclusively from the remaining 16 scenes using scripted pick-and-place policies. Figure~\ref{fig:libero_res} (bottom) plots policy success rate vs. number of collected trajectories. \textbf{Random} performs substantially worse than all other methods, while \textbf{VLM-Uniform} significantly better, demonstrating that VLM-based semantic priors can alone provide a strong signal for task generation. Incorporating principled data evaluation further improves performance: both \textbf{FSC} and \method{} outperform \textbf{STRAP} and no-evaluation baselines. Across all tasks and collection budgets, \method{} consistently matches or exceeds the strongest baseline, demonstrating the effectiveness of attribution-guided data collection.

\begin{figure*}[t]
    \centering
    \includegraphics[width=1.0\linewidth]{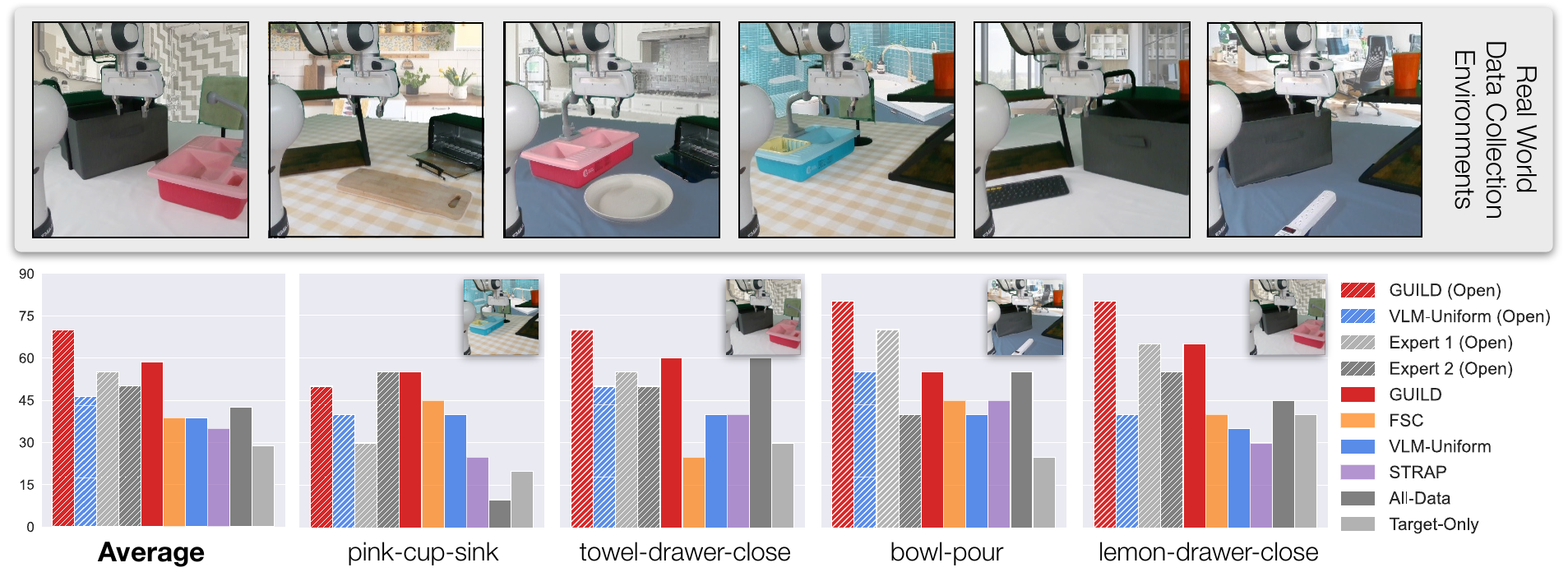}
    \caption{\textbf{Real world guided data collection.} We evaluate the policies trained on the data generated by different strategies. \method{} consistently achieves the highest success under the same data collection budget, highlighting its efficacy in guiding data collection towards useful tasks for the deployment setting.}
    \label{fig:real_res}
    \vspace{-1em}
\end{figure*}

We observe a similar trend in the real world, where we construct 6 environments spanning two layouts each of kitchens, bathrooms, and offices (Figure~\ref{fig:real_res}, top). To avoid recollecting data for every method, we first build an offline dataset containing 49 tasks across these environments~(Appendix~\ref{app:assets_real_closed}), which we refer to as the \emph{Closed-World} setting. Here, task selection is restricted to the existing dataset, and data collection corresponds to sampling trajectories from datasets of the selected tasks. We evaluate on 4 target tasks, holding out one environment as the deployment domain and using the remaining environments for data collection. Policies are trained on the selected datasets and evaluated over 20 rollouts. Results in Figure~\ref{fig:real_res} (solid bars) mirror simulation: \method{} achieves the best performance under the same collection budget. Notably, it also outperforms an \textbf{All-Data} baseline trained on the entire dataset, including data from the target environment, showing that targeted collection can be more effective than indiscriminately scaling dataset size.

\emph{Can \textbf{\method{}} guide data collection in open-world settings?}

Our primary goal is to deploy \method{} in \emph{Open-World} settings, where the system is not restricted to selecting from a fixed task pool, but can instead freely specify environments, object configurations, and task goals. We hypothesize that this additional flexibility amplifies the benefits of guided data collection by enabling more targeted exploration of the task space. Since evaluating multiple methods in this setting would require recollecting large real-world datasets, we compare only \method{} and \textbf{VLM-Uniform}, shown as the textured bars \includegraphics[scale=0.7]{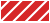} and \includegraphics[scale=0.7]{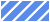} in Figure~\ref{fig:real_res}, respectively.
We observe two key trends. First, including data evaluation for GDC provides substantial gains over relying solely on VLM-generated task proposals: \method{} consistently outperforms \textbf{VLM-Uniform} across all evaluated tasks, improving average performance by approximately $25\%$. Second, both methods perform better in the \emph{Open-World} setting than in the corresponding \emph{Closed-World} setting, supporting our hypothesis that finer control over task specification leads to more effective data collection.

\emph{How does \textbf{\method{}} compare to human-guided data collection?}

\begin{wrapfigure}{r}{0.28\textwidth}
    \centering
    \vspace{-3.5em}

    \includegraphics[width=0.28\textwidth]{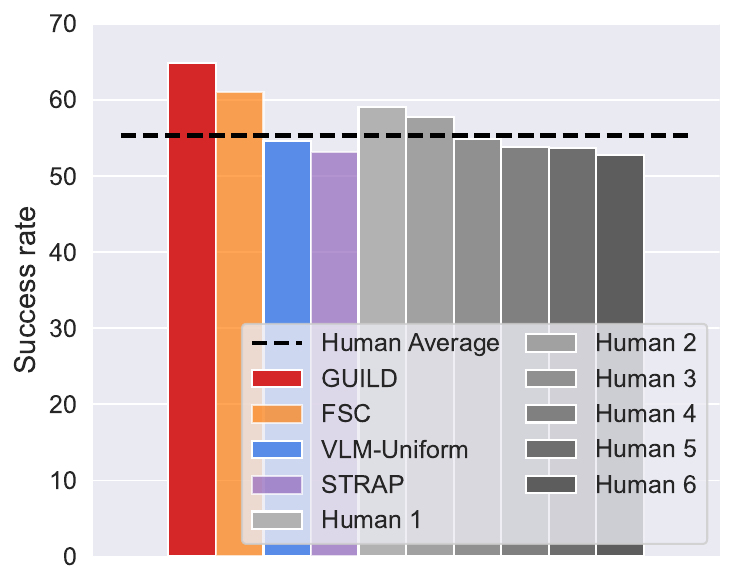}
    \caption{\textbf{Human Comparison.} Avg. performance on 3 LIBERO tasks of datasets collected via various methods and humans.}
    \label{fig:human_res}

    \vspace{-1em}
\end{wrapfigure}

Finally, we compare \method{} against humans, which remains the dominant paradigm for real-world robot data collection. In LIBERO, we survey 8 roboticist with varying levels of robotics experience (ranging from 1-5+ years), providing them the same information available to the GDC methods and asking them to propose tasks and allocate a fixed collection budget across them. Averaged results across the three tasks in LIBERO are shown in Figure~\ref{fig:human_res} (task-wise results and additional discussion provided in Appendix~\ref{app:task_wise_human}). While different humans have differing affinity for guiding data collection, the \textbf{Average Human} indicates the mean human performance, showing that on average, humans stay below the overall performance gains of \method{}. This suggests that attribution-based data evaluation provides a stronger signal for guiding collection of useful training data. Another interesting observation is that \textbf{VLM-Uniform} closely matches the performance of the \textbf{Average Human}, suggesting that modern VLMs capture much of the commonsense reasoning used by human experts and provide a strong prior for initializing the GDC process.

We observe a similar trend in the real world, where we compare against 2 human experts (4-5 years of robotics experience) in the \emph{Open-World} setting. Their individual performances are shown in Figure~\ref{fig:real_res} as \includegraphics[scale=0.7]{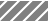} and \includegraphics[scale=0.7]{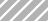}. Consistent with simulation, \method{} achieves higher average performance across tasks. We additionally observe that human-designed collection strategies are often inconsistent across tasks---eg. \textbf{Expert~1} on \texttt{towel-drawer-close} vs. \texttt{pink-cup-sink} task---whereas \method{} provides a more reliable and systematic mechanism for guiding data collection.

\subsection{Qualitative Analysis: What Data Does GUILD Collect?}
\label{sec:exp_qual_analysis}
\begin{figure}[t!]
    \centering
    \begin{subfigure}[b]{0.24\textwidth}
        \centering
        \includegraphics[width=1.0\linewidth]{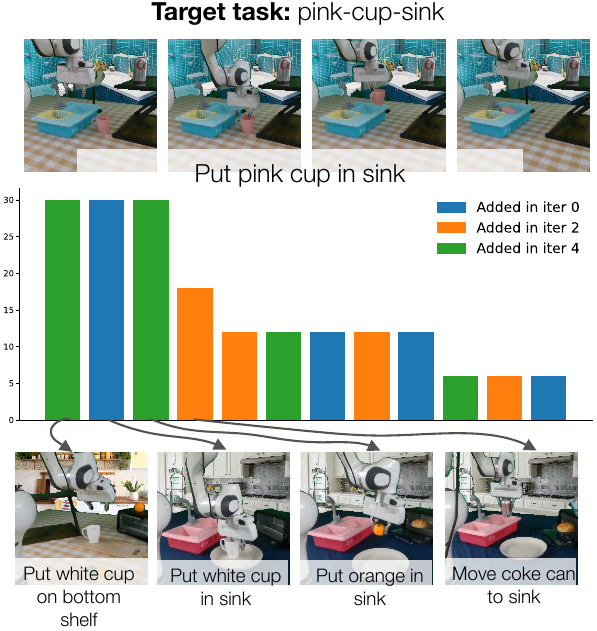}
    \end{subfigure}
    \hfill %
    \begin{subfigure}[b]{0.24\textwidth}
        \centering
        \includegraphics[width=1.0\linewidth]{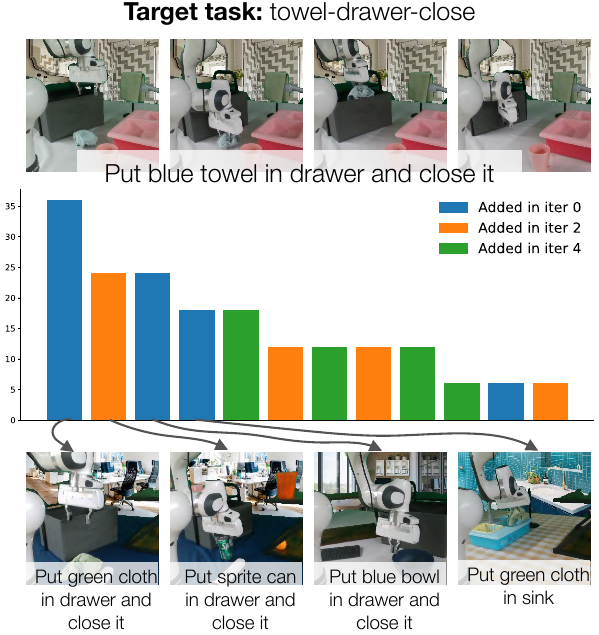}
    \end{subfigure}
    \hfill
    \begin{subfigure}[b]{0.24\textwidth}
        \centering
        \includegraphics[width=1.0\linewidth]{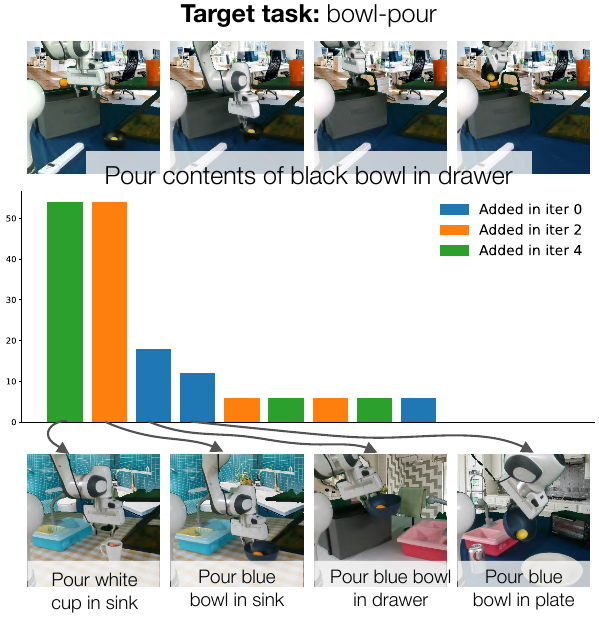}
    \end{subfigure}
    \hfill
    \begin{subfigure}[b]{0.24\textwidth}
        \centering
        \includegraphics[width=1.0\linewidth]{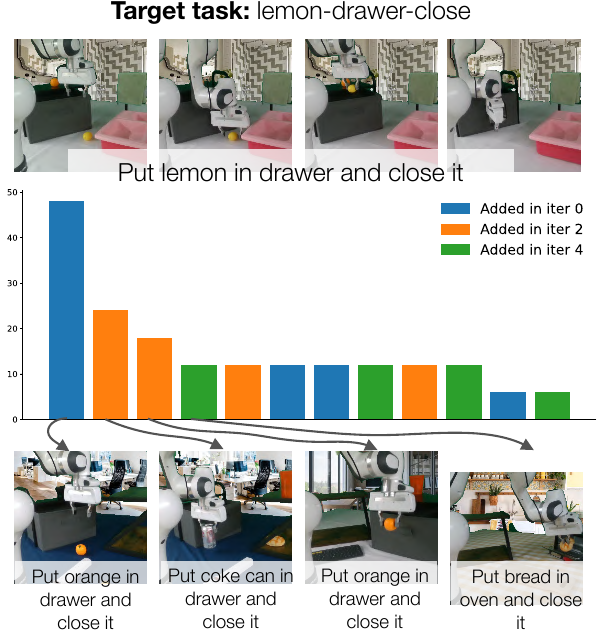}
    \end{subfigure}

    \caption{\textbf{Real-world qualitative analysis.} For each target task, we visualize the final allocation of collected trajectories across auxiliary tasks (\textbf{middle}), and tasks with the highest data allocation auxiliary tasks (\textbf{bottom}). Bar colors indicate the outer-loop iteration in which each auxiliary task was introduced.}
    \label{fig:real_qual_fig}
    \vspace{-1em}
\end{figure}

Figure~\ref{fig:real_qual_fig} visualizes the final data composition for each real-world target task collected by \method{} in the \emph{Open-World} setting. For each target, the auxiliary tasks are unique to that setting and are ordered from highest to lowest collected data. Each bar denotes the number of trajectories collected for an auxiliary task, while its color indicates the outer-loop iteration in which that task was introduced. We additionally visualize representative high-allocation auxiliary tasks for each target. We observe that \method{} prioritizes auxiliary tasks that preserve relevant semantic and behavioral structure with the target task, such as similar objects, receptacles, or manipulation patterns.

\paragraph{Useful tasks are discovered throughout the collection process.}
The color coding in Figure~\ref{fig:real_qual_fig} provides a view into when useful tasks are introduced during iterative data collection. Useful tasks are not necessarily contained in the initial proposal set: tasks introduced during later support-expansion rounds can still receive a substantial fraction of the final collection budget. For example, for \texttt{pink-cup-sink}, one of the highest-allocation tasks is introduced at a later support-expansion stage, while for \texttt{bowl-pour}, two of the largest allocations correspond to tasks introduced after initialization. This suggests that iterative support expansion is important: later proposals can introduce new high-utility tasks that were absent from the initial candidate set. Similar analysis in simulation mirrors this observation~(Appendix~\ref{app:sim_qual_results}).

\paragraph{Nature of the final data composition.}
We also observe that the shape of the final data distribution varies substantially across target tasks. For \texttt{bowl-pour}, and \texttt{lemon-drawer-close} the collection budget is relatively concentrated, with a small number of auxiliary tasks receiving a large fraction of the total allocation. In contrast, the final distributions for \texttt{pink-cup-sink}, and \texttt{towel-drawer-close} are more distributed, with substantial collection effort allocated across several auxiliary tasks rather than collapsing onto a single task. This illustrate that useful collection strategies can take qualitatively different forms depending on the target: in some cases, a small number of tasks dominate, while in others, performance benefits from a broader mixture of auxiliary tasks. This further motivates treating data collection as an adaptive allocation problem rather than relying on a fixed heuristic such as uniform sampling or selecting a single highly ranked task.

\section{Conclusion and Limitations}
\label{sec:conlusion}

We introduced \method{}, an iterative framework for Guided Data Collection that combines VLM-based task generation with predictive data attribution to actively decide what robot data to collect next. Across simulation and real-world experiments, \method{} substantially improves data efficiency over passive collection strategies, several baselines, and even experienced roboticists, highlighting the importance of treating data collection itself as an optimization problem. We conclude by highlighting a few limitations of \method{} and potential avenues of future work. First, while modern VLMs provide strong task priors, they can still generate unoriginal or infeasible tasks; in practice, these failures can often be mitigated through weak human supervision. Second, \method{} introduces a tradeoff between adaptivity and computational cost: smaller collection batches yield more accurate utility estimates but require more frequent attribution recomputation. A promising direction is to adaptively determine both batch sizes and attribution update schedules. Finally, our framework discretizes the task space through VLM-generated proposals before attribution-based evaluation. Extending attribution methods to continuous parameterizations could enable finer-grained exploration and broader application of such approaches.

\clearpage
\acknowledgments{This work was conducted in the Robot Interactive Intelligence (RobIn) Lab at The University of Texas at Austin and supported in part by the National Science Foundation (CAREER Award FAIN 2543745), the NVIDIA Academic Grant Program and TACC resources, and Amazon. Any opinions, findings, and conclusions or recommendations expressed in this material are those of the authors and do not necessarily reflect the views of the National Science Foundation or the other sponsors.}

\bibliography{example}  %
\clearpage
\newpage
\begin{appendices}
\renewcommand{\thesection}{\Alph{section}}

\section{Method Details}
\label{app:method}

\subsection{VLM Task Proposer}
\label{app:task_proposer_details}
Recall from Section~\ref{subsec:vlm} that \method{} uses a vision--language model (VLM) to propose task parameters ($\phi$) that are likely to yield useful training data for a target deployment task. The prompts used in each experimental setting are provided in Appendix~\ref{app:prompts}. Given a video and language description of the target task, together with the available training assets (e.g., scenes and manipulable objects), the VLM proposes candidate task parameters. Appendix~\ref{app:prompts} also includes the prompt used during \emph{feedback-driven support expansion}, where a ranked list of previously evaluated tasks is provided to guide future proposals. The corresponding inputs and outputs for each setting are provided in Appendix~\ref{app:assets}.

\textbf{Outer-Loop Schedule ($K$).}
As described in the main text, the VLM is periodically queried to expand the support of the collection distribution. Rather than using a fixed schedule, we generate tasks more aggressively during the early stages of Guided Data Collection and taper off later. Since the VLM defines the support over which evaluation-guided collection operates, early support expansion helps mitigate poor initial proposals and reduces the risk of allocating significant collection budget to uninformative tasks. In LIBERO, we perform $30$ iterations of Guided Data Collection and expand the task set after iterations $\{2,4,8,16\}$. In the real-world experiments, we perform $20$ iterations and expand the task set after iterations $\{3,9\}$.

\textbf{Explore-Exploit Task Generation.}
During initial experiments, we found that conditioning task generation solely on evaluation-ranked tasks (\emph{exploit}) often caused the VLM to repeatedly propose similar tasks, reducing diversity in the candidate pool. To mitigate this, we adopt an explore-exploit strategy inspired by reinforcement learning~\citep{sutton1998reinforcement}. At each support-expansion step, when proposing $m$ new tasks, a subset is generated using evaluation-ranked tasks (\emph{exploitation}), while the remainder is generated using only the original target-task context (\emph{exploration}), analogous to the initialization procedure. This strategy encourages diversity while still leveraging attribution feedback. In LIBERO, if $i$ denotes the outer-loop iteration (1-indexed), then $i$ of the $m$ proposed tasks are generated via exploitation and the remaining $m-i$ via exploration. Consequently, early iterations emphasize exploration when attribution estimates are noisy, while later iterations gradually shift toward exploitation as more reliable utility estimates become available. In real world, $i+1$ tasks were generated via \emph{exploitation} and $m-i-1$ via \emph{exploration}.

On the \texttt{Bowl-ketchup-drawer} task, using only \emph{exploitation} achieved $50.7\pm2.1\%$ average success over three independently collected datasets, whereas \emph{explore-exploit} improved performance to $56.9\pm1.8\%$. We did not observe significant differences on the remaining LIBERO tasks.

\subsection{Data Attribution -- Computing Metagradients}

As discussed in Section~\ref{subsec:attribution}, we use MAGIC~\citep{ilyas2025magic} to compute the attribution model. Eq.~\ref{eq:metagradient_approx} is reproduced below for convenience,
\[
\mathcal{M}(\mathcal{A}(\mathbf{w}))
\approx
\mathcal{M}(\mathcal{A}(\mathbf{w}_0))
+
\nabla_\mathbf{w}
\mathcal{M}(\mathcal{A}(\mathbf{w}_0))^\top
(\mathbf{w}-\mathbf{w}_0).
\]

There we discussed that computing $\mathcal{I}=\nabla_\mathbf{w}
\mathcal{M}(\mathcal{A}(\mathbf{w}_0))$ is the main challenge, requiring the computation of metagradients with respect to the data weights $\mathbf{w}$ which are the meta-parameters of the training process. Computing metagradients has traditionally been notoriously difficult, and consequently most prior works in data attribution have focused on approximating $\mathcal{I}$~\citep{koh2017understanding,park2023trak,bae2024training}. 

Recent advances, however, have shown that these metagradients can be computed both \emph{exactly} and \emph{efficiently} for modern deep learning systems~\citep{engstrom2025optimizing,calian2025datarater}. We use the REPLAY method of \citet{engstrom2025optimizing}, which computes the effect of each training trajectory on the final policy by differentiating through the policy training process itself.

Intuitively, training a policy can be viewed as a long sequence of optimization steps, where each gradient update is influenced by the training data. Since the final policy depends on all of these updates, measuring how a trajectory affects deployment performance requires tracing its influence through the entire training process. A naïve implementation would save the policy after every optimization step and then work backwards through this sequence of saved models to compute the metagradients. For modern policies, however, storing hundreds of intermediate models can require an impractical amount of memory. 

REPLAY~\citep{engstrom2025optimizing} avoids this issue by storing only a small number of checkpoints during training. Let $s_i$ denote a saved model checkpoint at iteration $i$. When an intermediate model at iteration $j$ is needed for metagradient computation, REPLAY reconstructs it by re-executing the optimization updates between the most recent saved checkpoint $s_i$ and iteration $j$, where $i<j$, rather than storing every intermediate model. As a result, exact metagradients can be computed with memory requirements being practical. As a result, we can efficiently compute the attribution $\mathcal{I}$ which quantifies how increasing or decreasing the weight of each trajectory would affect the target-domain evaluation metric.
We refer the reader to \citep{engstrom2025optimizing} and \citep{ilyas2025magic} for more details.

\subsubsection{Relationship between Utility and Downstream Performance}
\label{app:utility_corr}
\begin{wrapfigure}{r}{0.45\textwidth}
    \centering
    \vspace{-2em}

    \includegraphics[width=0.45\textwidth]{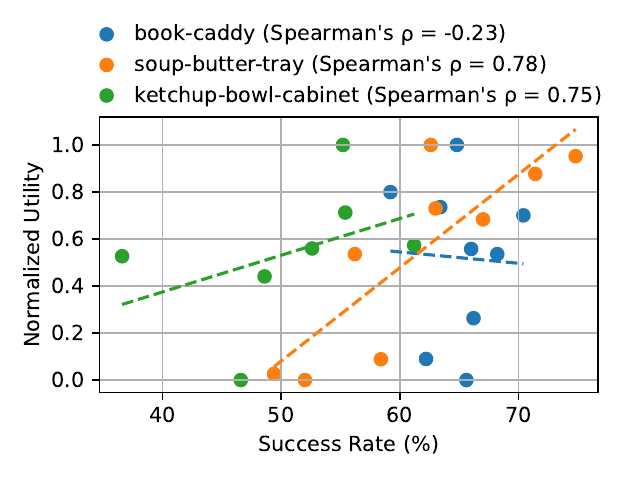}
    \caption{\textbf{Relationship between estimated utility and downstream policy performance.} Each point corresponds to a dataset collected by one of the evaluated data-collection strategies. We plot the dataset's mean task utility against the success rate of the policy trained on that dataset.}
    \label{fig:utility_corr}

    \vspace{-1em}
\end{wrapfigure}

A central assumption in \method{} is that the attribution-based marginal utility $U(\phi)$ provides a useful signal for deciding where to allocate additional data-collection effort. Intuitively, if $U(\phi)$ captures how beneficial data from task $\phi$ is for the target task, then datasets containing a larger proportion of high-utility data should tend to produce policies with stronger downstream performance. We therefore study whether the utility estimates produced by \method{} are associated with the eventual success of policies trained on different collected datasets.

To obtain datasets with different compositions, we use the datasets collected by the different data-collection baselines in our LIBERO experiments. For each dataset, we compute the utility of its constituent tasks and average these values to obtain a single dataset-level utility score. We then compare this score against the success rate of the policy trained on the corresponding dataset by computing Spearman's rank correlation for each task. Figure~\ref{fig:utility_corr} shows this relationship separately for each target task.

We observe a strong positive relationship for \texttt{ketchup-bowl-cabinet} and \texttt{soup-butter-tray}, with correlations of $\rho=0.75$ and $\rho=0.78$, respectively. In these settings, datasets assigned higher average utility by our attribution model also result in policies with higher downstream success. For \texttt{book-caddy}, the correlation is weakly negative ($-0.23$); notably, the evaluated datasets occupy a comparatively narrow range of success rates, making differences in downstream performance less pronounced. Overall, despite noise from stochastic policy evaluation and attribution estimation, these results provide evidence that the utility signal used by \method{} is meaningfully associated with downstream policy performance and can serve as a useful signal for guiding data collection.

\section{Training and Evaluation Details}
\label{app:train_and_eval}
\begin{table}[h]
\centering

\caption{\textbf{Task-specific hyperparameters for Guided Data Collection.} \underline{Data collection iterations} denote the total number of inner-loop collection iterations. \underline{Outer-loop schedule} specifies the inner-loop iterations at which support expansion is performed (Appendix~\ref{app:task_proposer_details}). \(m\) denotes the number of new task parameters proposed during each outer-loop iteration, and \underline{initial task demos} denotes the number of demonstrations collected when a new task is introduced. \underline{Batch size} (\(b\)) specifies the number of demonstrations collected during each inner-loop iteration using attribution-guided task selection.}

\resizebox{\columnwidth}{!}{
\begin{tabular}{lccccccc}
    \toprule
     \multirow{3}{*}{} &
     \multirow{3}{*}{No. of} &
     \multirow{3}{*}{No. of} &
     \multirow{3}{*}{No. of Data} &
     \multirow{3}{*}{Outer-loop} &
     \multirow{3}{*}{No. of tasks} &
     \multirow{3}{*}{Batch size} &
     \multirow{3}{*}{No. of Initial} \\
     & & & & & & \\
     & Training Envs &
       Target Demos &
       Collection Iter. &
       Schedule &
       Every Iter. ($m$) &
       ($b$) &
       Task Demos \\
    \midrule

    LIBERO & & & & & & \\
    \quad\texttt{Book-Caddy} & 16 & 3 & 30 & [2,4,8,16] & 5 & 20 & 5 \\
    \quad \texttt{Soup-butter-tray} & 16 & 5 & 30 & [2,4,8,16] & 5 & 20 & 5 \\
    \quad \texttt{Bowl-ketchup-drawer} & 16 & 5 & 30 & [2,4,8,16] & 5 & 10 & 5 \\

    \midrule

    Real World & & & & & & \\
    \quad \texttt{pink-cup-sink} & 5 & 15 & 20 & [3, 9] & 4 & 6 & 6\\
    \quad \texttt{towel-drawer-close} & 5 & 30 & 20 &  [3, 9] & 4 & 6 & 6 \\
    \quad \texttt{bowl-pour} & 5 & 20 & 20 &  [3, 9] & 3 & 6 & 6 \\
    \quad \texttt{lemon-drawer-close} & 5 & 50 & 20 & [3, 9] & 4 & 6 & 6 \\

    \bottomrule
\end{tabular}
}
\label{tab:gdc_hyperparams}
\end{table}

\subsection{LIBERO}

\textbf{Environment and Data Collection.}
Task-specific hyperparameters are reported in Table~\ref{tab:gdc_hyperparams}. For each target task, we use the remaining 16 environments from the LIBERO-90 suite as data collection environments. Demonstrations are generated using a scripted policy that takes a pick object and place region as input and outputs an expert trajectory.

\textbf{Task Parameterization ($\phi$).} In LIBERO, each target task is paired with 16 data-collection scenes selected from the LIBERO-90 suite. These scenes vary in background, table texture, and available manipulation objects. The input to \method{} and all baselines consists of this set of scenes together with the objects available in each scene (see Appendix~\ref{app:assets_libero}). The algorithm then proposes a set of tasks $\{\phi_i\}_{i=1}^{n}$, where each task is parameterized as
\[
\phi_i=(\mathcal{S},o,r).
\]
Here, $\mathcal{S}$ is a discrete index identifying the scene, $o$ denotes the object to manipulate, and $r$ denotes the target receptacle or placement region. We also allow $r=\emph{null}$, corresponding to a pick-only task involving object $o$. A summary of the parameterization is provided in Table~\ref{tab:task_parameterization}.

\textbf{Method Details.}
Following Algorithm~\ref{alg:method}, \method{} first uses the target-task context and available training environments to generate an initial set of $m$ candidate tasks. An initial set of demonstrations is generated for each proposed task (Table~\ref{tab:gdc_hyperparams} last column) using the scripted policy, after which the attribution module estimates a marginal utility score for every task. \method{} then allocates the next batch of the collection budget to the highest-utility task (which is generated by the scripted policy) and recomputes utilities on the updated dataset. This inner-loop process alternates between data evaluation and data collection. We perform a total of $30$ inner-loop iterations in the simulation tasks. Periodically, according to the outer-loop schedule in Table~\ref{tab:gdc_hyperparams}, the VLM expands the task set by proposing new tasks conditioned on the ranked list of previously evaluated tasks.

\textbf{Policy Training and Evaluation.}
After collecting a guided dataset $\mathcal{D}$, we fine-tune a language-conditioned Octo~\citep{octo_2023} policy initialized from the publicly released Octo-Small checkpoint. Following prior work, we co-train on the target dataset $\mathcal{D}_{\mathrm{targ}}$ and collected dataset $\mathcal{D}$ using a mixing ratio $\alpha=0.5$ for $10$k gradient steps. Policies are evaluated on the target task using 100 rollouts and results are averaged across 5 training seeds. We repeat the entire data collection and evaluation procedure with 3 different seeds; Figure~\ref{fig:libero_res} reports the mean across collection seeds, with error bars indicating standard error.

\begin{table}[h]
\centering
\vspace{-1em}

\caption{\textbf{Task-parameter spaces used across experimental settings.}
We summarize the components of the task parameter $\phi$ that may be varied by the task proposer in each setting. Scene selection and object configuration indicate whether these components are explicitly chosen during task generation, while the task specification describes how the manipulation objective is represented. The candidate-space column indicates whether proposals are restricted to a fixed task set or generated within a structured/open-ended space.}
\label{tab:task_parameterization}
\renewcommand{\arraystretch}{1.3}
\resizebox{\columnwidth}{!}{
\begin{tabular}{lcccc}
\toprule
\textbf{Setting} &
\textbf{Scene Selection} &
\textbf{Object Configuration} &
\textbf{Task Specification} &
\textbf{Candidate Space} \\
\midrule
\textbf{LIBERO} &
\yes &
\no &
Pick object / place region &
Structured \\

\textbf{Real-World} (Closed-World) &
\no &
\no &
Selection from provided task list &
Fixed \\

\textbf{Real-World} (Open-World) &
\yes &
\yes &
Natural-language instruction &
Open-ended \\
\bottomrule
\end{tabular}
}
\end{table}

\subsection{Real World}

\textbf{Environment and Data Collection.}
We construct six real-world environments, shown in Table~\ref{tab:real_open_assets}. To create distinct scenes, we vary appliance types and placements, table textures (using different tablecloths), and background appearances using a green screen that is replaced with scene-specific imagery during data collection and evaluation. Demonstrations are collected via human teleoperation~\citep{dass2024telemoma}. During data collection, for each trajectory, we place 1--3 distractor objects and randomly perturb object poses to increase scene diversity.

\textbf{Task Parameterization ($\phi$).} In the real-world \emph{Open-World} setting, each target task has five scenes available for data collection. The scenes vary in background and table texture, and each contains a fixed set of scene-specific objects that cannot be repositioned, representing realistic environments, such as kitchens, in which fixtures like sinks or dishwashers remain fixed. In addition, we provide a separate set of movable objects that can be placed within the scene (see Appendix~\ref{app:assets_real_open}). \method{} and all baselines take as input the available scenes and movable objects, and output a set of tasks $\{\phi_i\}_{i=1}^{n}$ parameterized as
\[
\phi_i=(\mathcal{S},\mathcal{C},\mathcal{T}).
\]
Here, $\mathcal{S}$ is a discrete index identifying the scene, $\mathcal{C}$ is a free-form language description of the scene configuration, specifying the relative placement of movable objects, e.g., \textcolor{softorange}{\emph{place the blue bowl with a lemon inside it on the table, centered between the pink sink and the white plate}} or \textcolor{softgreen}{\emph{keep the toaster oven door open and place the green towel on the table between the pink sink and the white plate}}. Together, $\mathcal{S}$ and $\mathcal{C}$ define the scene configuration. Finally, $\mathcal{T}$ is a free-form language instruction specifying the task to perform, e.g., \textcolor{softorange}{\emph{pour the contents of the blue bowl onto the white plate}} or \textcolor{softgreen}{\emph{put the green towel in the toaster oven and close the oven door}}.

In the real- \emph{Closed-World} setting, the candidate task set is fixed in advance~(Appendix~\ref{app:assets_real_closed}). The algorithms are given this set of tasks as input and output the indices of the tasks selected for data collection.

\textbf{Method Details.}
We evaluate \method{} in both \emph{Closed-World} and \emph{Open-World} settings.
In the \emph{Closed-World} setting, we first collect an offline dataset consisting of 49 tasks~(Appendix~\ref{app:assets}) distributed across the six environments. Here, the task space is fixed and data collection is simulated by selecting tasks from this dataset. The VLM task proposer is therefore restricted to choosing which existing tasks should receive additional collection budget. Following Algorithm~\ref{alg:method}, \method{} allocates demonstrations to tasks based on their estimated marginal utilities, periodically expanding the candidate task set from the pool of available tasks according to the outer-loop schedule in Table~\ref{tab:gdc_hyperparams}.

In the \emph{Open-World} setting, tasks are not restricted to a predefined dataset. Instead, the VLM proposes complete task specifications, including the scene, object configuration, and task description. These task specifications are then provided to a human teleoperator, who collects demonstrations accordingly. As in simulation, \method{} alternates between attribution-based utility estimation and data collection, while periodically expanding the candidate task set using feedback-driven support expansion.

\textbf{Policy Training and Evaluation.}
After obtaining a guided dataset $\mathcal{D}$, we fine-tune a language-conditioned Octo policy by co-training on $\mathcal{D}_{\mathrm{targ}}$ and $\mathcal{D}$ with a mixing ratio $\alpha=0.5$ for $25$k gradient steps. Policies are evaluated in the target environment using $20$ rollouts from a fixed set of $20$ initial object configurations shared across all methods to ensure fair comparison.

\subsection{Human Experiment Details}
To evaluate human-guided data collection, we surveyed participants with varying levels of robotics experience. The least experienced participant had approximately one year of robotics experience, while the most experienced participants had more than five years of research experience in robot learning and manipulation. To construct human-proposed datasets, we provided participants with the same information available to the automated methods, including the target-task context, descriptions of the available training environments, and the total data collection budget. Participants were instructed to design a data collection strategy that would maximize performance on the target task while respecting the specified budget constraints. Importantly, participants did not have access to attribution scores, baseline results, or any additional feedback beyond the initial information provided to the automated methods.

For each target task, participants proposed a set of tasks within the same parameter space used by \method{}. For example, in the \emph{Real Open-World} setting, each proposal specified a collection environment, object configuration, task description, and the amount of collection budget allocated to that task. The proposed datasets were then collected using the same data-collection process employed for the automated methods (scripted policies in simulation and human teleoperation in the real world). Finally, all human-guided datasets were evaluated using the identical policy training and evaluation protocol described for each of our settings. This ensures that differences in performance arise solely from the data collection strategy rather than the learning or evaluation procedure.

\section{Additional Results}
\label{app:additional_res}

\subsection{Real-World Qualitative Results}
\label{app:sim_qual_results}
\def\taskFigWidth{0.32\textwidth}
\def\taskFigWidthtwo{0.27\textwidth}

\begin{figure}
\captionsetup[subfigure]{justification=centering}
\centering
\begin{subfigure}[t]{\taskFigWidth}
    \centering
    \includegraphics[width=1.0\textwidth]{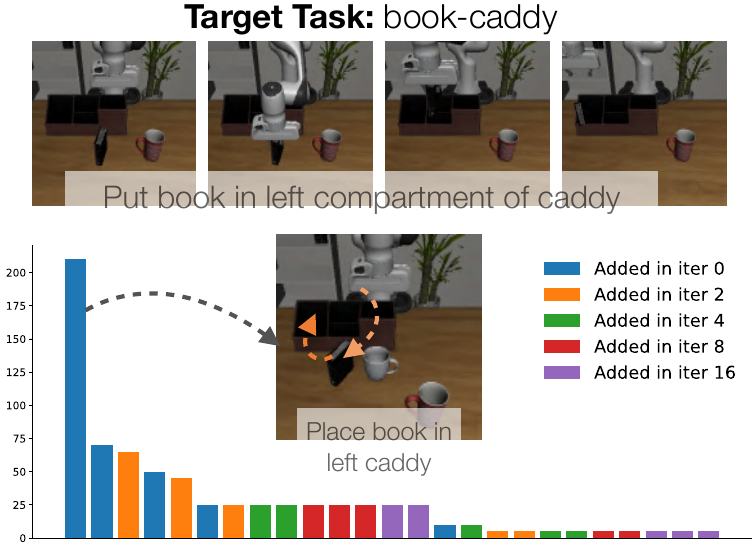}
\end{subfigure}
\begin{subfigure}[t]{\taskFigWidth}
    \centering
    \includegraphics[width=1.0\textwidth]{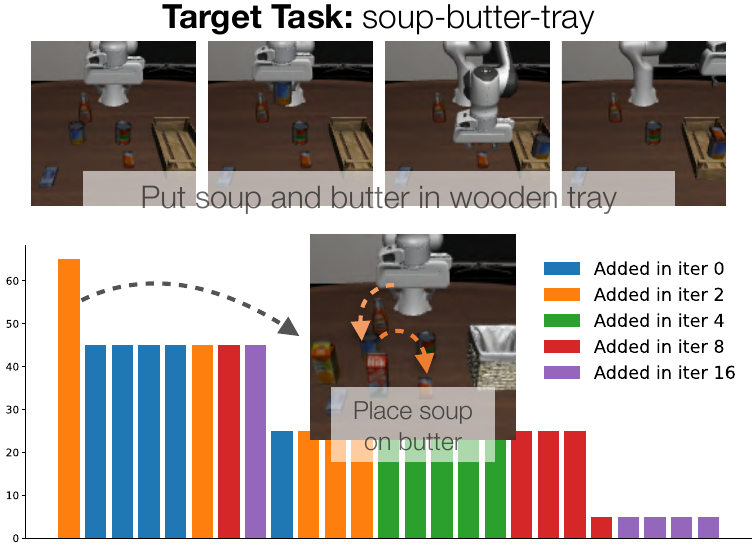}
\end{subfigure}
\begin{subfigure}[t]{\taskFigWidth}
    \centering
\includegraphics[width=1.0\textwidth]{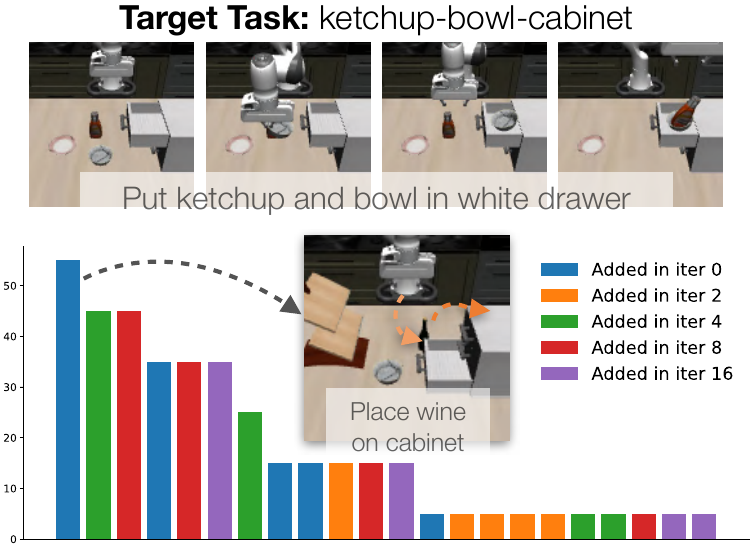}
\end{subfigure}
\caption{\textbf{Data Composition of data collected via GUILD.} We demonstrate the composition of the final datasets from each of the LIBERO tasks. \textbf{Top.} A visualization of the target task for which the data is collected. \textbf{Bottom.} Final data composition plot, where each bar shows the number of trajectories collected for a particular task.}
\label{fig:sim_qual_fig}
\end{figure}

In Figure~\ref{fig:sim_qual_fig}, we visualize the final data composition for each LIBERO target task for which we perform Guided Data Collection using \method{} and notice largely similar trends as we did in real world~(Section~\ref{sec:exp_qual_analysis}). We observe that the resulting mixtures are highly target-dependent and non-uniform. For \texttt{book-caddy}, the auxiliary task \emph{place book in left caddy} is extremely similar to the target task and receives by far the largest fraction of the collection budget, showing that \method{} can identify and exploit a particularly relevant source of data when one is available. In contrast, the final distributions for \texttt{soup-butter-tray} and \texttt{ketchup-bowl-cabinet} are more distributed, with substantial collection effort allocated across several auxiliary tasks rather than collapsing onto a single task. We also observe that useful tasks need not appear in the initial proposal set: for \texttt{soup-butter-tray}, the highest-allocation auxiliary task is introduced at a later support-expansion stage, while for \texttt{ketchup-bowl-cabinet}, several of the highest-allocation tasks are also proposed later in the iterative collection process. Together, these results illustrate that useful collection strategies can take different forms across targets and that iterative support expansion can uncover high-utility tasks beyond the initial candidate set.

\subsection{Human-Guided Data Collection Results}
\label{app:task_wise_human}
\begin{figure*}[t]
    \centering
    \includegraphics[width=1.0\linewidth]{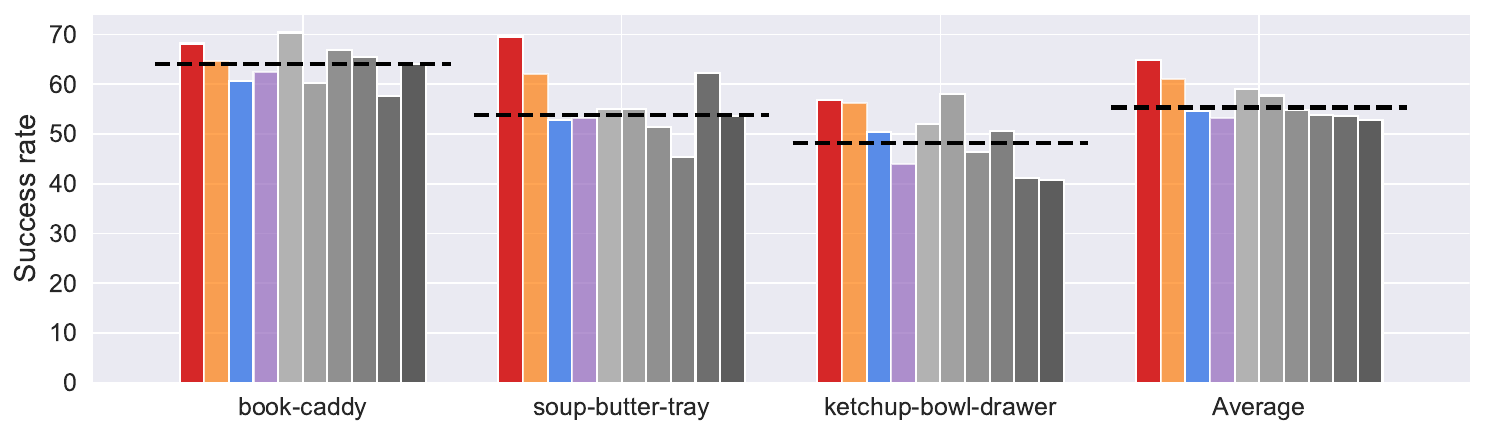}
    \caption{Task-wise human-guided data collection comparison on LIBERO}
    \label{fig:human_gen_libero_full}
\end{figure*}
We present the task-wise human-guided data collection results for LIBERO in Figure~\ref{fig:human_gen_libero_full}. In addition to our analysis in Section~\ref{sec:experiments}, we also observe substantial variability in human performance across tasks: participants who perform well on one task often do not achieve similar performance on others. This suggests that effective data collection requires reasoning about task transfer in ways that can be difficult to do consistently, even for experienced roboticists. In contrast, \method{} provides a systematic mechanism for evaluating data utility and achieves strong performance across all tasks. As a result, while individual experts may occasionally outperform \method{} on specific tasks, \method{} consistently exceeds the average expert performance and provides a more reliable strategy for guided data collection.

\subsection{Hard vs. Soft Sampling}
\label{app:soft_sampling}
In Section~\ref{subsec:attribution}, we described our attribution-guided collection strategy. Given task utilities $U(\phi)$, \method{} selects the highest-utility task parameters,
\[
\phi_j^*
\in
\arg\max_{\phi\in\mathcal{T}_i}
U(\phi),
\]
and asks a teleoperator to collect a batch of $b$ demonstrations from that task. We refer to this strategy as \emph{hard-sampling}, since all collection effort is allocated to the currently best task. An alternative approach is to convert task utilities into a sampling distribution and collect demonstrations probabilistically, which we refer to as \emph{soft-sampling}. Given utilities $U(\phi)$, we first discard tasks with negative utility (or positive utility when $\mathcal{M}$ is a minimizing objective like loss), ensuring that collection effort is not allocated to clearly harmful tasks. We then define a sampling distribution using a softmax:
\[
p(\phi)
=
\frac{\exp(U(\phi)/T)}
{\sum_{\phi' \in \mathcal{T}_i}
\exp(U(\phi')/T)},
\]
where $T$ is a temperature parameter. Demonstrations are then collected by sampling tasks according to $p(\phi)$. A comparison between hard- and soft-sampling in LIBERO is shown in Table~\ref{tab:ablations}, named \textbf{GUILD} and \textbf{GUILD}~(soft-sampling) respectively. 
We find that hard- and soft-sampling achieve broadly similar performance, suggesting that both are viable strategies for allocating collection effort. Soft-sampling provides a natural mechanism for interpolating between exploration and exploitation through the temperature parameter $T$. However, its performance is sensitive to the choice of $T$: larger values spread collection effort across many tasks, while smaller values resemble hard-sampling. In practice, selecting an appropriate temperature may require task-specific tuning, hence, we adopt hard-sampling throughout the paper due to its simplicity, robustness, and strong empirical performance.
\begin{table}[t]
\centering

\caption{Mean performance with standard error of policies trained on the final aggregated dataset by \textbf{\method{}}, \textbf{\method{}}~(soft-sampling), \textbf{\method{}}~(no-feedback) and \textbf{\method{}}~(gemini) for simulation tasks. The evaluation methodology is the same as our main experiments described in Appendix~\ref{app:train_and_eval}.}
\label{tab:ablations}
\begin{tabular}{lcccc}
\toprule
Tasks & book-caddy & soup-butter-tray & bowl-ketchup-drawer & \textbf{Average}\\

\midrule
\textbf{\method{}} & $68.2\pm1.3$ & $69.6\pm3.6$ & $56.9\pm1.8$ & $64.9$ \\
\textbf{\method{}} (soft-sampling) & $66.1\pm 0.5$ & $63.2\pm2.3$& $54.7\pm4.5$ & $61.3$\\
\textbf{\method{}} (no-feedback) & $65.5\pm 1.1$ & $55.3\pm3.9$& $56.9\pm1.7$ & $59.2$\\
\textbf{\method{}} (gemini) & $67.2\pm 2.0$ & $52.2\pm1.5$& $63.8\pm 0.6$ & $61.1$\\

\bottomrule
\end{tabular}

\end{table}

\subsection{Feedback based Support Expansion}
\label{app:no_feedback}
While our main experiments highlight the importance of data evaluation through attribution as a guidance signal for data collection, here we evaluate whether VLMs are able to interpret this signal and propose new tasks that are more relevant for the deployment task. As described in Section~\ref{subsec:vlm}, in each support-expansion step, we provide the VLM with a ranked list of previously proposed tasks, where the ranking is derived from the task utilities estimated during the data evaluation stage. This feedback allows the VLM to learn from earlier proposals and generate new task parameters that are more relevant to the target deployment task. We refer to this mechanism as \emph{feedback-driven support expansion}.

To evaluate the usefulness of this feedback, we ablate it in Table~\ref{tab:ablations}. \textbf{\method{}} denotes the full pipeline, while \textbf{\method{}}~(no-feedback) removes feedback-driven support expansion and instead generates new tasks using only the original target-task context. The results show that attribution-based feedback consistently matches or improves performance across all tasks. While the magnitude of the improvement varies, feedback is particularly beneficial on the \texttt{soup-butter-tray} task, where removing it reduces performance by approximately $15\%$. In contrast, both variants achieve similar performance on \texttt{bowl-ketchup-drawer}, suggesting that the VLM's commonsense priors alone are sufficient to discover useful tasks in this setting. Overall, these results indicate that VLMs can effectively incorporate attribution-based feedback, enabling more targeted exploration of the task space and improving the quality of future task proposals.

\subsection{Impact of the VLM Used}
\label{app}

Throughout our experiments, we use GPT-5.2~\citep{singh2025openai} for task proposal and feedback-driven support expansion. Here, we study the sensitivity of \method{} to the choice of VLM by replacing GPT-5.2 with Gemini-2.5-Flash~\citep{comanici2025gemini} in our LIBERO experiments, while keeping all other hyperparameters fixed. We report the results in Table~\ref{tab:ablations} as \textbf{GUILD}~(gemini).

We observe that the choice of VLM can substantially affect task-wise performance. On \texttt{soup-butter-tray}, Gemini achieves approximately $17\%$ lower success than GPT-5.2. In contrast, on \texttt{bowl-ketchup-drawer}, Gemini improves performance by approximately $7\%$, outperforming the best GPT-5.2-based variant evaluated for this task. These results suggest that different VLMs induce different task-proposal distributions, which can meaningfully affect the downstream data collected by \method{}. Nevertheless, averaged across the LIBERO tasks, the Gemini variant remains competitive with GPT-5.2, indicating that \method{} is not tied to a particular VLM. More broadly, improvements in the task-proposal model may provide an additional avenue for improving guided data collection. 

\subsection{Stage-wise results for Real world tasks}
\label{app:stage_wise_res}
\begin{figure*}[t]
    \centering
    \includegraphics[width=1.0\linewidth]{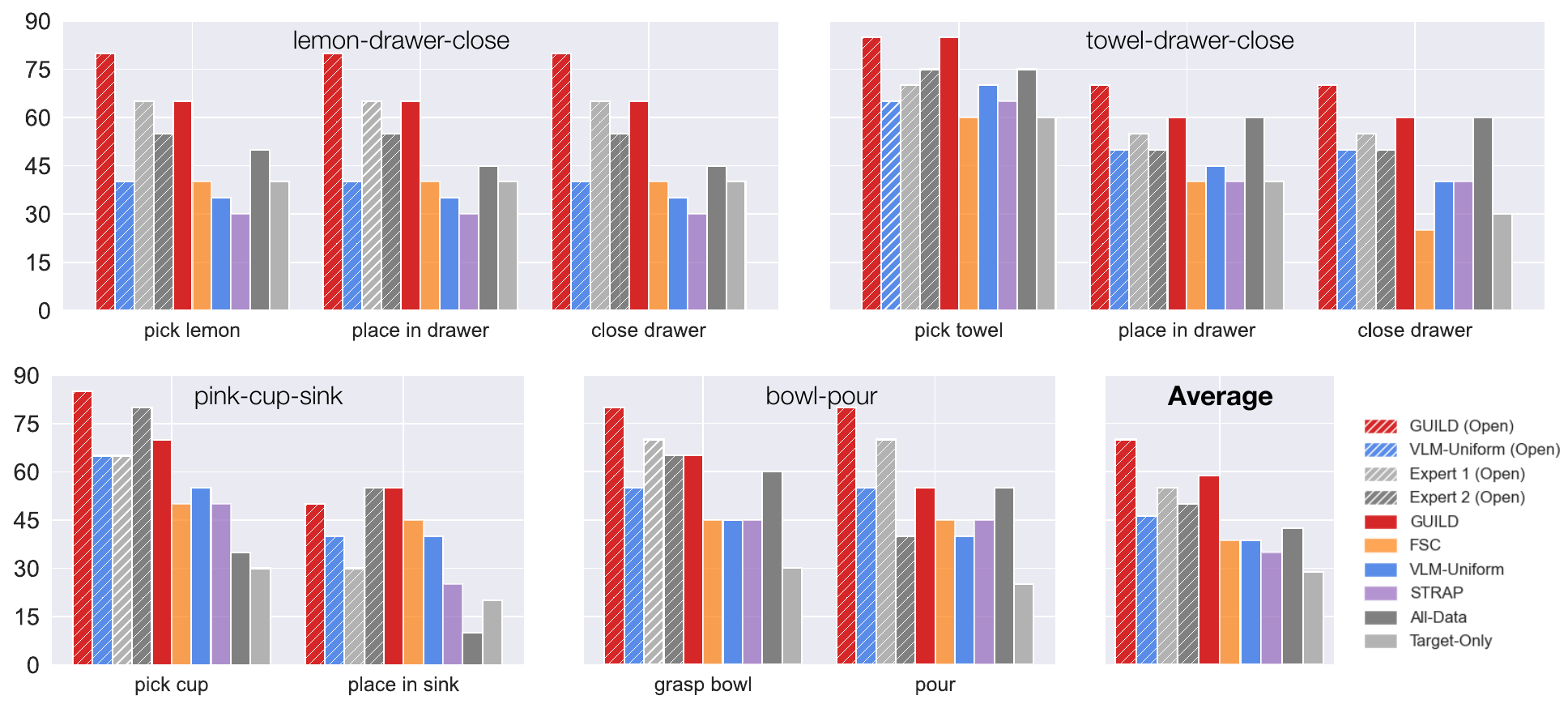}
    \caption{\textbf{Stage-wise real world results.} We perform 20 rollouts of a policy on a task and measure the progress of the policy at different stages. The success rates are absolute for each stage, counting it as a fail even if the policy fails in an earlier stage.}
    \label{fig:stage_wise_real}
    \vspace{-1em}
\end{figure*}

We measure stage-wise success rates for the real-world evaluations, shown in Figure~\ref{fig:stage_wise_real}. The stages for each task are listed below,
\begin{itemize}
    \item \texttt{pink-cup-sink}: pick pink cup $\rightarrow$ place in sink
    \item \texttt{towel-drawer-close}: pick towel $\rightarrow$ place it in the open drawer $\rightarrow$ close drawer
    \item \texttt{bowl-pour}: grasp black bowl $\rightarrow$ pour its contents in the drawer
    \item \texttt{lemon-drawer-close}: pick lemon $\rightarrow$ place it in the open drawer $\rightarrow$ close drawer
\end{itemize}
For \texttt{bowl-pour} and \texttt{lemon-drawer-close}, performance is largely determined by the first stage of the task: once the target object is successfully grasped, policies typically complete the remaining stages. In contrast, \texttt{pink-cup-sink} and \texttt{towel-drawer-close} require successful execution of later-stage interactions. For \texttt{pink-cup-sink}, failures primarily arise during placement, where policies may collide with the rim of the sink, drop the cup, or place it in the drying rack. For \texttt{towel-drawer-close}, the bottleneck is placing the deformable towel into the drawer; if the towel unfolds, policies often fail to recover before attempting to close the drawer. These results highlight that the evaluation suite spans a range of failure modes, from object acquisition to precise placement and deformable-object manipulation. Across all tasks, \method{} maintains strong performance throughout the task progression and consistently outperforms the baselines, suggesting that its gains are not tied to a particular setting.

\section{Baseline Implementation}
\label{app:baselines}

Implementation details for all baselines are provided below.

\begin{itemize}

\item \textbf{Random} and \textbf{VLM-Uniform.}
Both Random and VLM-Uniform are passive data collection strategies that uniformly allocate collection effort across a set of candidate tasks. We instantiate \textbf{Random} only in LIBERO, where task proposals are generated by randomly selecting a training environment, pick object, and placement region, while ensuring the resulting task is feasible. In contrast, \textbf{VLM-Uniform} uses the same VLM-based task proposer as \method{} to initialize the task set.
For both methods, no data evaluation is performed. During each collection iteration, demonstrations are requested from the task with the fewest collected trajectories (ties broken randomly), resulting in approximately uniform coverage of the task set. Comparing Random and VLM-Uniform isolates the value of commonsense-guided task initialization independent of attribution-based guidance.

\item \textbf{STRAP~\citep{memmel2025strap}.}
STRAP uses the same VLM-generated task proposals as \method{}, but replaces attribution-based evaluation with visual similarity estimation. We use the official implementation
(\url{https://github.com/WEIRDLabUW/STRAP}),
which embeds trajectory segments using DINOv2~\citep{oquab2023dinov2} and computes similarity to the target-task demonstrations using Dynamic Time Warping. The resulting similarity scores are normalized into a sampling distribution over tasks, with more similar tasks receiving a larger share of the collection budget. These scores also induce a ranking over tasks, which we use during feedback-driven support expansion (Section~\ref{subsec:vlm}) to guide future VLM task proposals.

\item \textbf{FSC~\citep{zha2025guiding}.}
Factored Scaling Curves (FSC) follows the same high-level pipeline as STRAP and \method{} but uses scaling-curve analysis to estimate task utility. For each task $\phi$, FSC samples $N$ datasets of increasing size from $\mathcal{D}_\phi$ ($N=4$ in all experiments), trains a policy on each subset, and evaluates its performance. The resulting performance-versus-data-size curve is used to estimate a scaling factor, which serves as the task utility. Tasks with larger scaling factors receive additional collection budget, and the induced ranking is used during support expansion.

FSC is considerably more expensive than \method{}. For $M$ tasks, each iteration requires training $M \times N$ policies, compared to a single policy for \method{}. To reduce this cost, we follow~\citet{zha2025guiding} and use the Group-$t$ variant, which partitions tasks into groups of size $t$. We use $t=5$ for LIBERO and $t=4$ for the real-world experiments (except \texttt{bowl-pour}, where $t=3$). Since online evaluation of every trained policy is prohibitively expensive, we adopt FSC's offline evaluation variant. Specifically, we measure cosine similarity between embeddings of the training and target datasets. In Octo~\citep{octo_2023}, this corresponds to the final transformer class token as the policy embedding. These modifications make FSC computationally feasible within our iterative Guided Data Collection framework.

\end{itemize}

\section{Metagradient Hyperparameters and Training Cost}
\label{app:meta_hparam_cost}

\begin{wraptable}{r}{8cm}
\vspace{-1em}
\centering
\caption{Hyperparameters for computing metagradients.}
\label{tab:metagradient_hparams}
\begin{tabular}{lcc}
\toprule
\textbf{Hyperparameter} & \textbf{LIBERO} & \textbf{Real-World}\\
\midrule
Policy architecture & Octo-Small & Octo-Small \\
Warmup steps $N_w$ & Adaptive & Adaptive \\
Metagradient steps $N_m$ & Adaptive & Adaptive \\
Max step bound $C$ & 500 & 300 \\
Batch size $B$ & 512 & 512 \\
Optimizer & Adam & Adam \\
Learning-rate scheduler & Cosine & Cosine \\
\bottomrule
\end{tabular}
\vspace{-1em}
\end{wraptable}

At each iteration of Guided Data Collection, we estimate the metagradient $\mathcal{I}$ for each trajectory in the current training dataset and aggregate these scores to compute task-level marginal utilities. The hyperparameters used for this computation are summarized in Table~\ref{tab:metagradient_hparams}.

Following~\citet{engstrom2025optimizing}, each metagradient computation consists of a warmup phase of $N_w$ optimization steps followed by a metagradient phase of $N_m$ steps. Since the dataset grows throughout the collection process, we adapt both quantities based on the number of optimization steps in one epoch,
\[
E=\left\lfloor\frac{|\mathcal{D}|}{B}\right\rfloor.
\]
where $\mathcal{D}$ is the dataset and $\mathcal{B}$ is the batch size. We first choose the metagradient horizon based on dataset size: $N_m=50$ when $E<50$, $N_m=75$ when $50\leq E<75$, and $N_m=100$ otherwise. We then set the warmup length as
\[
N_w=\min(C-N_m,\,10E),
\]
where $C$ is the maximum number of policy-training steps. If $N_m < E$, which happens in certain cases, we set $N_m=E$ ensuring that the metagradient phase spans at least one full epoch. Thus, for small datasets we differentiate through multiple epochs, while for larger datasets the metagradient horizon grows with the dataset size. The warmup phase scales up to approximately ten epochs, subject to the overall policy-training budget.

\paragraph{Training Cost.}
Each metagradient computation and utility-estimation step takes approximately 10--15 minutes on four NVIDIA A5000 GPUs. In comparison, data collection itself takes only a few minutes per iteration in simulation and approximately 15--30 minutes per iteration in the real world, depending on the task. Including attribution and data collection together, the complete GUILD loop takes approximately 7--8 hours for 30 LIBERO iterations and 10--15 hours for 20 real-world open-world iterations.

\section{Prompts used for VLM task-proposer}
\label{app:prompts}
We provide the prompts used for task proposals for each of our settings below. In each prompt the common elements are \textbf{black}, the elements that appear only in initializing prompt are \textbf{\pone{blue}} and the elements that only appear in \emph{feedback-driven support expansion} are \textbf{\ptwo{red}}. The input assets for each of the settings are provided in Appendix~\ref{app:assets}.

\subsection{LIBERO}

\begin{promptbox}{LIBERO Combined Task Proposal Prompt}
You are helping generate training data for robot learning for a given DEPLOYMENT TASK. The DEPLOYMENT TASK is provided below and you need to create similar tasks in training environments.

\medskip
DEPLOYMENT TASK Description: \textit{$<$target task description$>$}\\
DEPLOYMENT TASK video frames: \textit{[Frame images 1/6 -- 6/6]}

\medskip
Your goal is to generate training scenarios that are similar to the DEPLOYMENT TASK but adapted to the training environment and available objects.

\medskip
Following are the TRAINING SCENES:

\medskip
TRAINING SCENE 1 (ID: 1)\\
Available pick objects: \textit{$<$list of pick objects for SCENE 1$>$}\\
Available place regions: \textit{$<$list of place regions for SCENE 1$>$}\\
Visual example (first frame): \textit{[Image]}

\medskip
TRAINING SCENE 2 (ID: 2)\\
Available pick objects: \textit{$<$list of pick objects for SCENE 2$>$}\\
Available place regions: \textit{$<$list of place regions for SCENE 2$>$}\\
Visual example (first frame): \textit{[Image]}

\medskip
$\vdots$

\medskip
\ptwo{Based on previous training results, here is a RANKING of how well previously selected tasks helped improve the performance of the deployment task -- ordered from best (Rank 1) to worst (Rank \textit{M}). Use this ranking to guide your selection of new training scenarios but DON'T repeat any of the existing ranked tasks in your proposals.}

\medskip
\ptwo{RANKING OF PREVIOUS TRAINING TASKS:}

\medskip
\ptwo{RANK 1/\textit{M}:\\
Scene ID: \textit{$<$scene id$>$}\\
Task Description: \textit{$<$task description$>$}}

\medskip
\ptwo{RANK 2/\textit{M}:\\
Scene ID: \textit{$<$scene id$>$}\\
Task Description: \textit{$<$task description$>$}}

\medskip
\ptwo{$\vdots$}

\medskip
Please generate \textit{M} different training scenarios that capture the essence of the deployment task but are adapted to the training scene.

\medskip
For each scenario, provide:
\begin{enumerate}
    \item Scene ID: The scene where the given scenario should take place
    \item Task Description: Describe a pick or pick-place task involving an object and a region
\end{enumerate}
Format your response exactly as follows:

\medskip
SCENARIO 1:\\
SCENE ID: [SELECTED SCENE ID]\\
TASK DESCRIPTION: [Pick $<$obj1$>$ / Pick $<$obj1$>$ and place in $<$obj2$>$]\\
REASON: [brief explanation]

\medskip
SCENARIO 2:\\
SCENE ID: [SELECTED SCENE ID]\\
TASK DESCRIPTION: [Pick $<$obj1$>$ / Pick $<$obj1$>$ and place in $<$obj2$>$]\\
REASON: [brief explanation]

\medskip
Continue for all \textit{N} scenarios.

\medskip
Guidelines:
\begin{itemize}
    \item \pone{Use objects available in the selected scene that are similar to those in deployment}
    \item \pone{Keep the core manipulation skill/behavior similar to the deployment task}
    \item Only use the OBJECTS/REGIONS available in the selected scene
    \item Ensure tasks are feasible given the training scene setup
    \item Make sure the task is a single PICK or PICK-PLACE task only
    \item \pone{Try to suggest diverse training scenarios}
    \item \ptwo{Prefer training scenarios similar (but not repeating) to higher-ranked tasks, as they have proven more effective in improving deployment performance}
    \item \ptwo{IMPORTANT: DON'T REPEAT any of the existing RANKED TASKS in your proposals. Propose new tasks, using similar objects in different scenes or different objects in the same scene.}
    \item Ensure that TASK DESCRIPTION uses only objects chosen from the available objects in the selected training scene
\end{itemize}
\end{promptbox}

\subsection{Real (Open-World)}
\begin{promptbox}{Real (Open) World Combined Task Proposal Prompt}
You are helping generate training data for robot learning for a given DEPLOYMENT TASK. The DEPLOYMENT TASK is provided below and you need to create similar tasks in training environments.

\medskip
DEPLOYMENT TASK Description: \textit{$<$target task description$>$}\\
DEPLOYMENT TASK video frames: \textit{[Frame images 1/5 -- 5/5]}

\medskip
Your goal is to generate training scenarios that are similar to the DEPLOYMENT TASK but adapted to the training environment and available objects.

\medskip
TRAINING ENVIRONMENTS:\\
- Available objects: \textit{$<$list of available objects$>$}\\
- Following are the TRAINING SCENES:

\medskip
TRAINING SCENE 1 (ID: 1)\\
Visual example: \textit{[Image]}

\medskip
TRAINING SCENE 2 (ID: 2)\\
Visual example: \textit{[Image]}

\medskip
$\vdots$

\medskip
\ptwo{Based on previous training results, here is a RANKING of how well previously selected tasks helped improve the performance of the deployment task -- ordered from best (Rank 1) to worst (Rank \textit{M}). Use this ranking to guide your selection of new training scenarios but DON'T repeat any of the existing ranked tasks in your proposals.}

\medskip
\ptwo{RANKING OF PREVIOUS TRAINING TASKS:}

\medskip
\ptwo{RANK 1/\textit{M}:\\
Scene ID: \textit{$<$scene id$>$}}\\
\ptwo{Task Description: \textit{$<$task description$>$}\\
Object Placement: \textit{$<$object placement$>$}}

\medskip
\ptwo{RANK 2/\textit{M}:\\
Scene ID: \textit{$<$scene id$>$}\\
Task Description: \textit{$<$task description$>$}\\
Object Placement: \textit{$<$object placement$>$}}

\medskip
\ptwo{$\vdots$}

\medskip
Please generate \textit{N} different training scenarios that capture the essence of the deployment task but are adapted to the training environment and available objects.

\medskip
For each scenario, provide:
\begin{enumerate}
    \item Scene ID: Which training scene to use (from the list of training scenes provided)
    \item Object Placement: Which objects from the available list to place and where in the training scene
    \item Task Description: What task the robot should perform (similar to deployment but adapted)
    \item Reason: A brief explanation of why this scenario is relevant for training the deployment task
\end{enumerate}
Format your response exactly as follows:

\medskip
SCENARIO 1:\\
Scene ID: [ID of the selected training scene]\\
Object Placement: [specific description of object placement in training scene]\\
Task Description: [task description adapted for training environment]\\
Reason: [brief explanation of relevance for training deployment task]

\medskip
SCENARIO 2:\\
Scene ID: [ID of the selected training scene]\\
Object Placement: [specific description of object placement in training scene]\\
Task Description: [task description adapted for training environment]\\
Reason: [brief explanation of relevance for training deployment task]

\medskip
And so on... Continue for all $N$ scenarios.

\medskip
Here are some EXAMPLES of how to adapt deployment tasks to training scenarios based on the training scenes and available objects. Use these examples as a guide for how to format your response and construct the sentence structure of the suggestions, while keeping the core manipulation skill/behavior similar to the deployment task.

\medskip
=======EXAMPLES START=======

Object Placement: Place the bread on the cutting board\\
Task Description: Put the bread in the oven

\medskip
Object Placement: Place the green towel on the table between cutting board and oven\\
Task Description: Drape green towel on the top of the wooden shelf

\medskip
Object Placement: Place the blue bowl on the top of the wooden shelf\\
Task Description: Put blue bowl in the bottom region of the wooden shelf

\medskip
Object Placement: Keep drawer closed and place white cup in the middle of the table\\
Task Description: Open drawer and place white cup in it

\medskip
Object Placement: Keep drawer open and place the sprite can on the right side of the table\\
Task Description: Put sprite can in drawer and close it

\medskip
Object Placement: Place the white cup in the oven and fill it with some colored paper balls\\
Task Description: Pour the contents of white cup into the sink

\medskip
Object Placement: Place the black bowl in the sink and leave the drawer open\\
Task Description: Put black bowl in drawer

\medskip
Object Placement: Place the green towel in the middle of the table (right side of the blue sink)\\
Task Description: Hang the green towel on the towel hanger

=======EXAMPLES END=======

\medskip
Guidelines:
\begin{itemize}
    \item Try to use objects from the available list that are similar to those in deployment
    \item Adapt the task to work with the training scene layout and available objects
    \item Make sure to not change the object placements of objects that already exist in the training scene
    \item Keep the core manipulation skill/behavior similar to the deployment task
    \item Align the sentence structure of the object placement and task description with the examples provided above
    \item Ensure tasks are feasible given the training scene setup
    \item \pone{IMPORTANT: suggest diverse training scenarios involving a variety of scenes, objects and skills, that may not immediately seem useful but cover useful behaviors}
    \item \ptwo{IMPORTANT: DON'T REPEAT any of the existing ranked tasks in your proposals. Propose new tasks using similar objects in different scenes or different objects in the same scene}
\end{itemize}

\medskip
Before proposing any new tasks, for each scene, list out the items you can identify. This will help ensure you are aware of the available items in each scene and can refer to them in your proposals.
\end{promptbox}

\subsection{Real (Closed-World)}
\begin{promptbox}{Real (Closed) World Combined Task Proposal Prompt}
You are helping generate training data for robot learning for a given DEPLOYMENT TASK. The DEPLOYMENT TASK is provided below and you need to create similar tasks in training environments.

\medskip
DEPLOYMENT TASK Description: \textit{$<$target task description$>$}\\
DEPLOYMENT TASK video frames: \textit{[Frame images 1/5 -- 5/5]}

\medskip
Your goal is to select the top-\textit{N} TRAINING TASKS that would be most beneficial for learning the DEPLOYMENT TASK from the provided list below. Consider the following properties when picking tasks:
\begin{enumerate}
    \item Similar manipulation skills/behaviors
    \item Similar object interactions
    \item Similar task structure or sub-goals
    \item Transferable skills that could help with the deployment task
\end{enumerate}
TRAINING TASK OPTIONS:

\medskip
TRAINING TASK 1 (ID: 1):\\
Task Description: \textit{$<$task description$>$}\\
Visual example (first frame): \textit{[Image]}

\medskip
TRAINING TASK 2 (ID: 2):\\
Task Description: \textit{$<$task description$>$}\\
Visual example (first frame): \textit{[Image]}

\medskip
$\vdots$

\medskip
\ptwo{Based on previous training results, here is a RANKING of how well previously selected tasks helped improve the performance of the deployment task -- ordered from best (Rank 1) to worst} 
\ptwo{(Rank \textit{M}). Use this ranking to guide your selection of new training tasks but DON'T repeat any of the existing ranked tasks in your proposals.}

\medskip
\ptwo{RANKING OF PREVIOUS TRAINING TASKS:}

\medskip
\ptwo{PREVIOUS TRAINING TASK 1 (RANK 1/\textit{M}):\\
Task Description: \textit{$<$task description$>$}\\
Visual example (first frame): \textit{[Image]}}

\medskip
\ptwo{PREVIOUS TRAINING TASK 2 (RANK 2/\textit{M}):\\
Task Description: \textit{$<$task description$>$}\\
Visual example (first frame): \textit{[Image]}}

\medskip
\ptwo{$\vdots$}

\medskip
Please select the top-\textit{N} training tasks that would be most beneficial for learning the deployment task. Provide your reasoning for each selection and give them in the ranking order of most to least beneficial.

\medskip
Format your response exactly as follows:

\medskip
SELECTED TASKS:\\
1. Task ID: [ID] - Reason: [brief explanation]\\
2. Task ID: [ID] - Reason: [brief explanation]\\
\vdots

\textit{M}. Task ID: [ID] - Reason: [brief explanation]

\medskip
Make sure to only select 3 unique tasks (no duplicates) and use the exact Task ID numbers from above.

\medskip
Guidelines:
\begin{itemize}
    \item Propose tasks that could help improve the performance of the DEPLOYMENT TASK
    \item Pay attention to the scene layout, objects in the scene and how those would affect the robot's motion
    \item Try to suggest diverse training tasks
    \item Use the exact Task ID numbers from the TRAINING TASK list when selecting tasks
    \item \ptwo{Prefer training scenarios that are similar to higher-ranked tasks, as they have proven more effective in improving deployment performance}
    \item \ptwo{DON'T REPEAT any of the existing ranked tasks in your proposals}
\end{itemize}
\end{promptbox}

\section{Training Environment Assets}
\label{app:assets}

\subsection{LIBERO}
\label{app:assets_libero}
\def\imagewidth{1.5cm}

\begin{longtable}{>{\centering\arraybackslash}m{0.15\textwidth} >{\raggedright\arraybackslash}m{0.25\textwidth} >{\raggedright\arraybackslash}m{0.5\textwidth}}
    \toprule
    \textbf{Scene Image} & \textbf{Pick Objects} & \textbf{Place Objects} \\
    \midrule
    \endfirsthead
    \toprule
    \textbf{Scene Image} & \textbf{Pick Objects} & \textbf{Place Objects} \\
    \midrule
    \endhead
    \bottomrule
    \endfoot

    \includegraphics[width=\imagewidth]{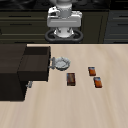} & akita\_black\_bowl\_1, butter\_1, butter\_2, chocolate\_pudding\_1 & wooden\_cabinet\_1\_top\_side, wooden\_cabinet\_1\_top\_region, akita\_black\_bowl\_1, butter\_1, butter\_2, chocolate\_pudding\_1 \\ \midrule

    \includegraphics[width=\imagewidth]{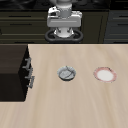} & akita\_black\_bowl\_1, plate\_1 & wooden\_cabinet\_1\_top\_side, wooden\_cabinet\_1\_top\_region, akita\_black\_bowl\_1, plate\_1 \\ \midrule

    \includegraphics[width=\imagewidth]{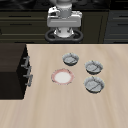} & akita\_black\_bowl\_1, akita\_black\_bowl\_2, akita\_black\_bowl\_3, plate\_1 & wooden\_cabinet\_1\_top\_side, wooden\_cabinet\_1\_top\_region, akita\_black\_bowl\_1, akita\_black\_bowl\_2, akita\_black\_bowl\_3, plate\_1 \\ 
    \midrule

    \includegraphics[width=\imagewidth]{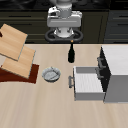} & akita\_black\_bowl\_1, wine\_bottle\_1 & white\_cabinet\_1\_top\_side, white\_cabinet\_1\_top\_region, white\_cabinet\_1\_bottom\_region, wine\_rack\_1\_top\_region, akita\_black\_bowl\_1, wine\_bottle\_1 \\ 
    \midrule

    \includegraphics[width=\imagewidth]{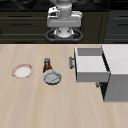} & akita\_black\_bowl\_1, plate\_1, ketchup\_1 & white\_cabinet\_1\_top\_side, white\_cabinet\_1\_top\_region, white\_cabinet\_1\_bottom\_region, akita\_black\_bowl\_1, plate\_1, ketchup\_1 \\ 
    \midrule

    \includegraphics[width=\imagewidth]{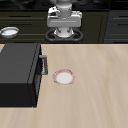} & white\_bowl\_1, plate\_1 & microwave\_1\_top\_side, plate\_1 \\ \midrule

    \includegraphics[width=\imagewidth]{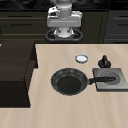} & white\_bowl\_1 & flat\_stove\_1\_cook\_region, wooden\_two\_layer\_shelf\_1\_top\_side, chefmate\_8\_frypan\_1 \\ \midrule

    \includegraphics[width=\imagewidth]{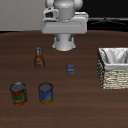} & alphabet\_soup\_1, cream\_cheese\_1, tomato\_sauce\_1, ketchup\_1 & basket\_1\_contain\_region, alphabet\_soup\_1, cream\_cheese\_1, tomato\_sauce\_1, ketchup\_1 \\ \midrule

    \includegraphics[width=\imagewidth]{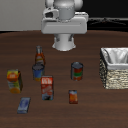} & alphabet\_soup\_1, cream\_cheese\_1, tomato\_sauce\_1, ketchup\_1, orange\_juice\_1, milk\_1, butter\_1 & basket\_1\_contain\_region, alphabet\_soup\_1, cream\_cheese\_1, tomato\_sauce\_1, ketchup\_1, orange\_juice\_1, milk\_1, butter\_1 \\ \midrule

    \includegraphics[width=\imagewidth]{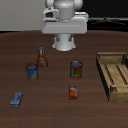} & alphabet\_soup\_1, cream\_cheese\_1, tomato\_sauce\_1, ketchup\_1, butter\_1 & wooden\_tray\_1\_contain\_region, alphabet\_soup\_1, cream\_cheese\_1, tomato\_sauce\_1, ketchup\_1, butter\_1 \\ \midrule

    \includegraphics[width=\imagewidth]{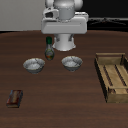} & akita\_black\_bowl\_1, akita\_black\_bowl\_2, chocolate\_pudding\_1 & wooden\_tray\_1\_contain\_region, akita\_black\_bowl\_1, akita\_black\_bowl\_2, new\_salad\_dressing\_1, chocolate\_pudding\_1 \\ \midrule

    \includegraphics[width=\imagewidth]{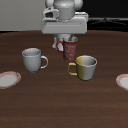} & porcelain\_mug\_1, red\_coffee\_mug\_1, white\_yellow\_mug\_1, plate\_1, plate\_2 & living\_room\_table\_plate\_left\_region, living\_room\_table\_plate\_right\_region, porcelain\_mug\_1, red\_coffee\_mug\_1, white\_yellow\_mug\_1, plate\_1, plate\_2 \\ \midrule

    \includegraphics[width=\imagewidth]{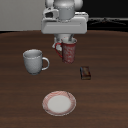} & porcelain\_mug\_1, red\_coffee\_mug\_1, plate\_1, chocolate\_pudding\_1 & living\_room\_table\_plate\_left\_region, living\_room\_table\_plate\_right\_region, porcelain\_mug\_1, red\_coffee\_mug\_1, plate\_1, chocolate\_pudding\_1 \\ \midrule

    \includegraphics[width=\imagewidth]{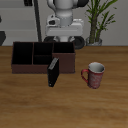} & black\_book\_1, red\_coffee\_mug\_1 & desk\_caddy\_1\_right\_contain\_region, desk\_caddy\_1\_left\_contain\_region, black\_book\_1, red\_coffee\_mug\_1 \\ \midrule

    \includegraphics[width=\imagewidth]{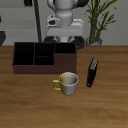} & black\_book\_1, white\_yellow\_mug\_1 & study\_table\_desk\_caddy\_right\_region, desk\_caddy\_1\_right\_contain\_region, desk\_caddy\_1\_left\_contain\_region, black\_book\_1, white\_yellow\_mug\_1 \\ \midrule

    \includegraphics[width=\imagewidth]{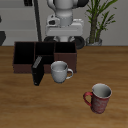} & black\_book\_1, red\_coffee\_mug\_1, porcelain\_mug\_1 & study\_table\_desk\_caddy\_right\_region, desk\_caddy\_1\_right\_contain\_region, desk\_caddy\_1\_left\_contain\_region, black\_book\_1, red\_coffee\_mug\_1, porcelain\_mug\_1 \\ \midrule

    \includegraphics[width=\imagewidth]{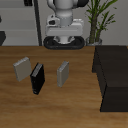} & black\_book\_1, yellow\_book\_1, yellow\_book\_2 & wooden\_two\_layer\_shelf\_1\_top\_side, black\_book\_1, yellow\_book\_1, yellow\_book\_2 \\ 

\end{longtable}

\subsection{Real (Open-World)}
\label{app:assets_real_open}
In the Real Open-World setting, the VLM task proposer is provided with the set of available scenes listed in Table~\ref{tab:real_open_assets}, together with the following movable objects: \textit{black bowl}, \textit{blue bowl}, \textit{blue towel}, \textit{bread}, \textit{coke can}, \textit{green towel}, \textit{lemon}, \textit{orange}, \textit{pink cup}, \textit{sprite can}, and \textit{white cup}. Given these assets, the VLM proposes task parameters specifying the collection environment, object configuration, and task description. We exclude both the target scene and target object from the assets available to the VLM. For example, when evaluating the \texttt{lemon-drawer-close} task in scene \textbf{b1}, the scene \textbf{b1} and the object \textit{lemon} are removed from the lists of available training scenes and objects, respectively.

\def\imagewidth{0.15\columnwidth}

\begin{table}[h]
\centering
\resizebox{1.\columnwidth}{!}{
\begin{tabular}{cccccc}
\toprule
Kitchen 1 (\textbf{k1}) & Kitchen 2 (\textbf{k2}) & Bathroom 1 (\textbf{b1}) & Bathroom 2 (\textbf{b2}) & Office 1 (\textbf{o1}) & Office 2 (\textbf{o2})\\
\includegraphics[width=\imagewidth]{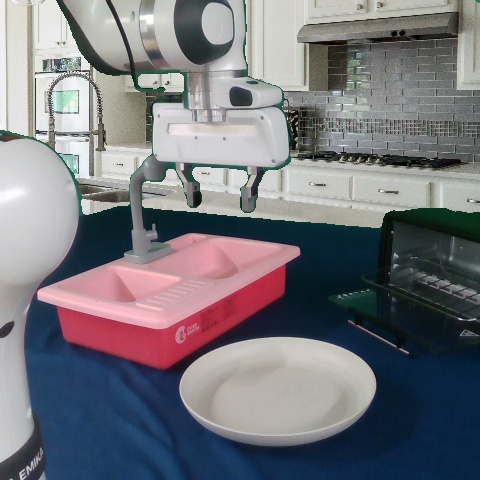} & \includegraphics[width=\imagewidth]{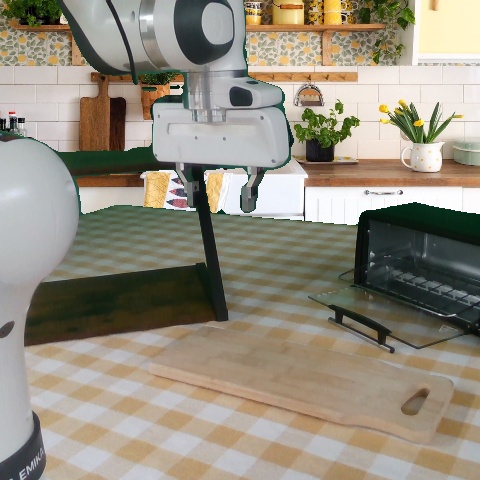} & \includegraphics[width=\imagewidth]{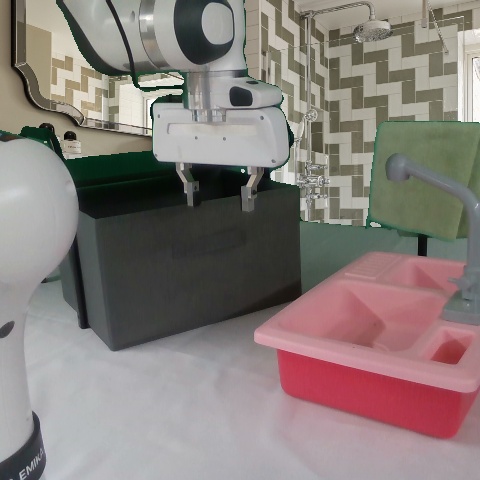} & \includegraphics[width=\imagewidth]{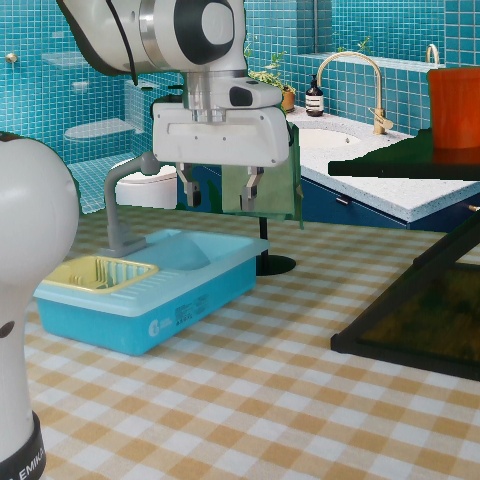} & 
\includegraphics[width=\imagewidth]{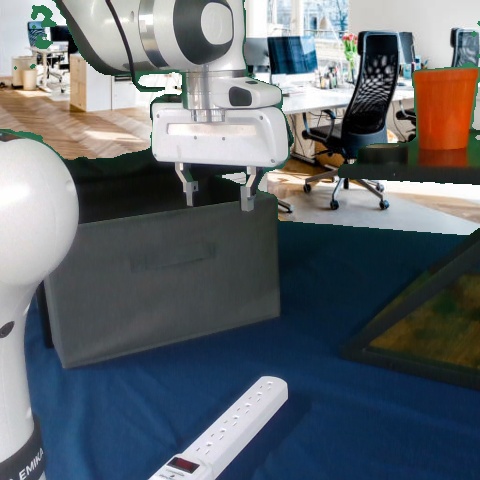} &
\includegraphics[width=\imagewidth]{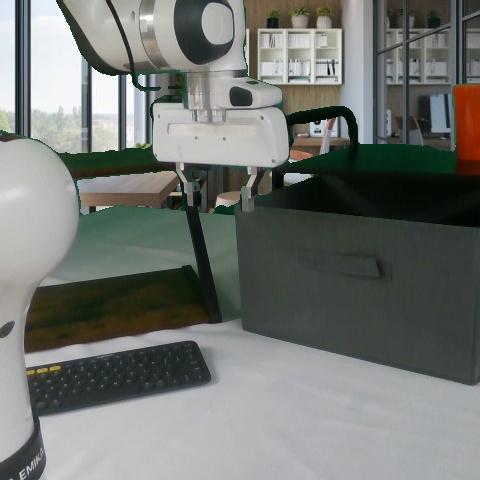} \\
\bottomrule
\end{tabular}
}
\vspace{0.5em}
\caption{Real world scene list}
\label{tab:real_open_assets}
\vspace{-2em}

\end{table}

\subsection{Real (Closed-World)}
\label{app:assets_real_closed}
In the Real Closed-World setting, we pre-collect a dataset containing 49 diverse tasks, listed below. The Scene IDs correspond to the environments shown in Table~\ref{tab:real_open_assets}. For a target task in a particular scene, all tasks from that scene are excluded from the Guided Data Collection process and may not be selected by the task proposer. List of all tasks is provided in the table below.

The \textbf{All-Data} baseline is trained on the target-task demonstrations together with the entire pre-collected dataset, including tasks from the target scene. This differs from all other methods, which are restricted from using data originating from the target environment.

\begingroup
\renewcommand{\arraystretch}{1.4}
\begin{longtable}{>{\centering\arraybackslash}p{0.1\textwidth} >{\raggedright\arraybackslash}p{0.4\textwidth} >{\raggedright\arraybackslash}p{0.4\textwidth}}
    \toprule
    \textbf{Scene ID} & \textbf{Initial Object Placement} & \textbf{Task} \\
    \midrule
    \endfirsthead
    \toprule
    \textbf{Scene ID} & \textbf{Initial Object Placement} & \textbf{Task} \\
    \midrule
    \endhead
    \bottomrule
    \endfoot

    k1 & White cup was placed on the plate & Put white cup in the sink \\
    k1 & Bread was placed on the plate & Put bread in the oven \\
    k1 & Black bowl was placed in the oven & Put black bowl in the sink \\
    k1 & Orange was placed on the plate & Put orange in the oven \\
    k1 & Coke can was placed in the drying rack & Put coke can on the plate \\
    k1 & White cup was placed in the oven and filled with some colored paper balls & Pour the contents of white cup into the sink \\
    k1 & Black bowl was placed on the plate and blue bowl was placed in the drying rack & Stack the black bowl on blue bowl \\
    k1 & Blue towel was placed in the sink & Put the blue towel next to plate \\
    k1 & Orange was placed in the oven & Put orange on plate \\

    k2 & Bread was placed on the cutting board & Put the bread in the oven \\
    k2 & Lemon was placed on the cutting board & Put the lemon in the oven \\
    k2 & Blue bowl was placed in the oven & Put blue bowl in the bottom region of the shelf \\
    k2 & Coke can was placed on the cutting board & Put coke can in the bottom region of the shelf \\
    k2 & Sprite can was placed in the bottom region of the shelf & Put sprite can on the cutting board \\
    k2 & Green towel was placed on the table between cutting board and oven & Drape green towel on the top shelf \\
    k2 & Bread was placed on cutting board and blue bowl was placed in the middle of the table & Put the bread in blue bowl \\
    k2 & Pink cup was placed on the cutting board & Put pink cup on the bottom shelf \\
    k2 & Lemon was placed in the oven & Put lemon on cutting board \\

    b1 & Drawer was left slightly open & Open the drawer \\
    b1 & Drawer was left open (not fully) & Close the drawer \\
    b1 & Blue towel was placed in the middle of the table (left side of the sink) & Hang the blue towel on the towel hanger \\
    b1 & Pink cup was placed in the middle of the table & Put pink cup in the sink \\
    b1 & Black bowl was placed in the sink and drawer was left open & Put black bowl in drawer \\
    b1 & Black bowl was placed in the middle of the table and filled with colored paper balls & Pour the contents of black bowl in the sink \\
    b1 & Drawer was left open (not fully) and pink cup was placed in the middle of the table & Put pink cup in drawer and close it \\
    b1 & Lemon was placed on top of the drawer and drawer was left slightly open & Slide the lemon into the drawer \\

    b2 & Green towel was placed in the middle of the table (right side of the blue sink) & Hang the green towel on the towel hanger \\
    b2 & Green towel was placed in the middle of the table (right side of the blue sink) & Put green towel on top of the shelf \\
    b2 & Coke can was placed on the right side of the table (before the shelf) & Put coke can on top of the shelf \\
    b2 & Pink cup was placed in the middle of the table & Put pink cup in the sink \\
    b2 & Black bowl was placed in the middle of the table and filled with colored paper balls & Pour the contents of black bowl in sink \\
    b2 & Black bowl was placed in the middle of the table and blue bowl was placed in the drying rack & Stack the black bowl on blue bowl \\
    b2 & Blue bowl was placed on the right side of the table and orange was placed on top of the shelf & Put orange in blue bowl \\
    b2 & Green towel was hung on the towel hanger & Slide the green towel in sink \\

    o1 & Drawer was left slightly open & Open the drawer \\
    o1 & Drawer was left open (not fully) & Close the drawer \\
    o1 & Sprite can was placed on the right side of the table (before the shelf) & Put sprite can on top of the shelf \\
    o1 & Blue towel was placed on the right side of the table (before the shelf) & Put blue towel in the drawer \\
    o1 & White cup was placed in the middle of the table and drawer was left open & Put white cup in the drawer and close it \\
    o1 & Orange was kept on top of the shelf and drawer was left open & Put orange in drawer and close \\
    o1 & Drawer was left slightly open and white cup was placed in the middle of the table & Open drawer and place white cup in it \\
    o1 & Both bowls were placed next to each other in the middle of the table & Stack black bowl on blue bowl \\
    o1 & Black bowl was placed in the middle of the table and drawer was left open & Put black bowl in drawer and close it \\

    o2 & Drawer was left slightly open & Open the drawer \\
    o2 & Drawer was left open (not fully) & Close the drawer \\
    o2 & Coke can was placed on the right side of the table (before the drawer) & Put the coke can in the bottom region of the shelf \\
    o2 & Pink cup was placed in the middle of the table and drawer was left open & Put pink cup in drawer and close it \\
    o2 & Green towel was placed on the top shelf & Put green towel in the drawer \\
    o2 & Lemon was kept on the bottom shelf and drawer was left open & Put lemon in drawer and close it \\
    o2 & Drawer was left slightly open and pink cup was placed in the middle of the table & Open drawer and place pink cup in it \\
    o2 & Black bowl was kept on the bottom shelf and drawer was left open & Black bowl in drawer and close it \\

\end{longtable}
\endgroup

\end{appendices}

\end{document}